\newcommand{\longversion}[1]{}
\newcommand{\shortversion}[1]{#1}
\newcommand{\springerversion}[1]{}
\newcommand{\arxivversion}[1]{#1}

\arxivversion{
 \documentclass[10pt,letter]{article}%
  \usepackage[hscale=0.7,scale=0.75]{geometry}}
\springerversion{\documentclass[runningheads,nvcountsame,a4paper,USenglish]{llncs}}

\usepackage{multirow}
\usepackage{amssymb,amsfonts,amsmath}%
\usepackage{xspace}
\usepackage[ruled,vlined,linesnumbered]{algorithm2e}
\usepackage[table,x11names]{xcolor}
\usepackage{graphicx}
\renewcommand*{\algorithmcfname}{Listing}
\SetKwInput{KwData}{In}
\SetKwInput{KwResult}{Out}
\SetAlFnt{\small}
\SetAlCapFnt{\small}
\SetAlCapNameFnt{\small}
\SetAlCapHSkip{0pt}
\SetEndCharOfAlgoLine{}

\usepackage{microtype}
\usepackage{tikz}
\usetikzlibrary{shapes}
\usetikzlibrary{calc}
\usetikzlibrary{positioning}
\tikzstyle{tdnode} = [draw,rounded corners,top color=vertexTopColor,bottom color=vertexBottomColor,minimum size=1.5em]
\tikzstyle{stdnode} = [tdnode, font=\scriptsize]
\tikzstyle{stdnodecompact} = [stdnode, inner sep = 1.5pt, outer sep = 0.1pt]
\tikzstyle{stdnodetable} = [stdnode, inner sep = 1.5pt, outer sep = 0]
\tikzstyle{stdnodenum} = [minimum size=1.5em, font=\scriptsize]
\tikzstyle{tdedge} = [-,draw,thick]
\tikzstyle{tdlabel} = [draw=none, rectangle, fill=none, inner sep=0pt, font=\scriptsize]
\colorlet{vertexTopColor}{white}
\colorlet{vertexBottomColor}{black!10}

\usepackage{ctable}
\usepackage{ifthen}
\newif\iflong
\usepackage{todonotes}

\usepackage{float}

\newcommand{\PMC}{\textsc{PMC}\xspace}%
\newcommand{\AlgA}{\algo{A}}%
\newcommand{\PROJ}{\algo{PROJ}\xspace}
\newcommand{\var}{\text{\normalfont var}}
\newcommand{\bigO}[1]{\ensuremath{{\mathcal O}(#1)}}

\newcommand{\Prev}[0]{{\it{PP}}\hy Tabs\xspace}

\DeclareMathOperator{\bucket}{=_P}%
\DeclareMathOperator{\buckets}{buckets}
\DeclareMathOperator{\subbuckets}{sub\hy buckets}
\DeclareMathOperator{\children}{children}
\DeclareMathOperator{\pcnt}{pmc}
\DeclareMathOperator{\sipmc}{s-ipmc}
\DeclareMathOperator{\orig}{\algo{SAT}\hy origins}
\DeclareMathOperator{\origs}{\algo{SAT}\hy origins}
\newcommand{\origse}[1]{\operatorname{#1\hy origins}}

\DeclareMathOperator{\local}{local}

\DeclareMathOperator{\Ext}{Ext}
\DeclareMathOperator{\Exts}{Exts}
\DeclareMathOperator{\PExt}{SatExt}

\DeclareMathOperator{\pmc}{pmc}
\DeclareMathOperator{\ipmc}{ipmc}

\DeclareMathOperator{\icnt}{ipmc}

\newcommand{\Tab}[1]{\ensuremath{\text{Child-Tabs}}}

\def\thyph{\text{-}\penalty0\hskip0pt\relax}
\newcommand{\ATabs}[2]{\ensuremath{#1\thyph\text{Comp[$#2$]}}}
\newcommand{\ATab}[1]{\ensuremath{#1\thyph\text{Comp}}}

\springerversion{\spnewtheorem{observation}{Observation}{\bfseries}{\itshape}}
\springerversion{\renewenvironment{example}{\begin{EXa}}{\hfill\ensuremath{\blacksquare}\end{EXa}}
\spnewtheorem{EXa}{Example}{\bfseries}{\normalfont}

\renewenvironment{proof}{\begin{pf}}{\qed\end{pf}}}

\arxivversion{\usepackage{amsthm}
\newtheorem{observation}{Observation}
\newtheorem{example}{Example}
\newtheorem{definition}{Definition}
\newtheorem{theorem}{Theorem}
\newtheorem{remark}{Remark}
\newtheorem{corollary}{Corollary}
\newtheorem{proposition}{Proposition}
\newtheorem{lemma}{Lemma}

}

\newenvironment{restateobservation}[1][\unskip]{%
  \begingroup
  \def\theobservation{\ref{#1}} 
}%
{%
  \addtocounter{observation}{-1}
  \endgroup
}%

\newenvironment{restatecorollary}[1][\unskip]{%
  \begingroup
  \def\thecorollary{\ref{#1}} 
}%
{%
  \addtocounter{corollary}{-1}
  \endgroup
}%

\newenvironment{restatetheorem}[1][\unskip]{%
  \begingroup
  \def\thetheorem{\ref{#1}} 
}%
{%
  \addtocounter{theorem}{-1}
  \endgroup
}%

\makeatletter
\newcommand{\algorithmfootnote}[2][\footnotesize]{
  \let\old@algocf@finish\@algocf@finish
  \def\@algocf@finish{\old@algocf@finish
    \leavevmode\rlap{\begin{minipage}{\linewidth}
    #1#2
    \end{minipage}}
  }
}
\makeatother

\graphicspath{{./figures/}}
\DeclareGraphicsExtensions{.pdf,.png,.jpg}
\arxivversion{\bibliographystyle{abbrv}\usepackage{hyperref}}
\springerversion{\bibliographystyle{splncs03}
\toctitle{Lecture Notes in Computer Science}
\tocauthor{Authors' Instructions}}

\title{Exploiting Treewidth for Projected Model Counting and its
  Limits%
\thanks{The work has been supported by the Austrian Science Fund
  (FWF), Grants Y698 and P26696, and the German Science Fund (DFG),
  Grant HO 1294/11-1. The first two authors are also affiliated with
  the 
  University of Potsdam, Germany.  The final publication will be available at Springer proceedings of SAT 2018.
}
}
\usepackage{url}
\urldef{\mailsa}\path|{fichte, hecher,morak,woltran}@dbai.tuwien.ac.at|
\springerversion{\titlerunning{Exploiting Treewidth for Projected Model Counting \& Limits}}

\author{Johannes K. Fichte\arxivversion{$^1$}%
\and Markus Hecher\arxivversion{$^1$}\and Michael Morak\arxivversion{$^1$}\and Stefan Woltran\arxivversion{$^1$}%
\arxivversion{\\[3pt]
    $^1$: TU Wien, Austria, \mailsa\\}
}

\springerversion{\institute{Institute of Logic and Computation, TU Wien, Vienna, Austria\\
  \email{lastname@dbai.tuwien.ac.at}\\
}
\authorrunning{Fichte et al.}}

\newenvironment{indented}{\begin{changemargin}{1cm}{0cm}}{\end{changemargin}}

\let\phi\varphi
\let\epsilon\varepsilon

\renewcommand{\models}{\vDash}

\newcommand{\ta}[1]{\ensuremath{2^{#1}}}
\newcommand{\card}[1]{\left|#1\right|}
\newcommand{\CCard}[1]{\|#1\|}

\newcommand{\Nat}{\mathbb{N}} 

\newcommand{\algo}[1]{\ensuremath{\mathbb{#1}}}

\newcommand{\NP}{\ensuremath{\textsc{NP}}\xspace}

\newcommand{\PSPACE}{\ensuremath{\textsc{PSpace}}}

\newcommand{\tw}[1]{\mathit{tw}(#1)}

\newcommand{\SB}{\{}%
\newcommand{\SM}{\mid}%
\newcommand{\SE}{\}}%
\def\hy{\hbox{-}\nobreak\hskip0pt}

\newcommand{\solver}[1]{\mbox{\text{#1}}\xspace}

\newcommand{\dynasp}[1]{\ensuremath{\solver{DynASP2}}}
\newcommand{\dynaspplus}[1]{\ensuremath{\solver{DynASP2.5}}}
\newcommand{\prog}{\ensuremath{F}}

\DeclareMathOperator{\width}{width}

\newcommand{\ASP}{\textsc{ASP}\xspace}
\newcommand{\cSAT}{\textsc{\#Sat}\xspace}

\makeatletter
\DeclareRobustCommand{\rvdots}{%
  \vbox{
    \baselineskip4\p@\lineskiplimit\z@
    \kern-\p@
    \hbox{.}\hbox{.}\hbox{.}
  }}
\makeatother

\DeclareFontFamily{U}{matha}{\hyphenchar\font45}
\DeclareFontShape{U}{matha}{m}{n}{
      <5> <6> <7> <8> <9> <10> gen * matha
      <10.95> matha10 <12> <14.4> <17.28> <20.74> <24.88> matha12
      }{}
\DeclareSymbolFont{matha}{U}{matha}{m}{n}
\DeclareMathSymbol{\squplus}{2}{matha}{"5D}

\makeatletter

\newcommand{\raisemath}[1]{\mathpalette{\raisem@th{#1}}}
\newcommand{\raisem@th}[3]{\raisebox{#1}{$#2#3$}}

\newcommand{\pushright}[1]{\ifmeasuring@#1\else\omit\hfill\ensuremath{\displaystyle#1}\fi\ignorespaces}
\newcommand{\pushleft}[1]{\ifmeasuring@#1\else\omit$\displaystyle#1$\hfill\fi\ignorespaces}
\providecommand{\leftsquigarrow}{%
	\mathrel{\mathpalette\reflect@squig\relax}%
}
\newcommand{\reflect@squig}[2]{%
	\reflectbox{$\m@th#1\rightsquigarrow$}%
}

\makeatother

\DeclareMathOperator{\cntc}{\#\cdot}%

\DeclareMathOperator{\type}{type}
\newcommand{\intr}{\textit{int}}
\newcommand{\leaf}{\textit{leaf}}
\newcommand{\rem}{\textit{rem}}
\newcommand{\join}{\textit{join}}

\let\P\undefined
\DeclareMathOperator{\P}{P}

\newcommand{\BIGOP}[1]
{
\mathop{\mathchoice%
{\raise-0.22em\hbox{\huge $#1$}}%
{\raise-0.05em\hbox{\Large $#1$}}{\hbox{\large $#1$}}{#1}}}

\newcommand{\BIGboxplus}{\mathop{\mathchoice%
{\raise-0.35em\hbox{\huge $\boxplus$}}%
{\raise-0.15em\hbox{\Large $\boxplus$}}{\hbox{\large $\boxplus$}}{\boxplus}}}

\newcommand{\TTT}{\ensuremath{\mathcal{T}}}%
\newcommand{\WWW}{\ensuremath{\mathcal{W}}}%

\newcommand*\mcupinn[2]{\vcenter{\hbox{$\mathsurround=0pt
  \ifx\displaystyle#1\textstyle\else#1\fi\bigcup$}}}

\newcommand{\NAT}{\ensuremath{\mathbb{N}}}

\newcommand{\inputPredColor}{orange!55!red}
\newcommand{\outputPredColor}{blue!45!black}
\newcommand{\statePredColor}{green!62!black}

\newcommand{\tuplecolor}[1]{\textcolor{#1}}

\newcommand{\tabval}{\ensuremath{u}}
\newcommand{\tab}[1]{\ensuremath{\tau_{#1}}}

\newcommand{\progt}[1]{\ensuremath{\prog_{\hspace{-0.05em}\leq\hspace{-0.05em}#1}}}

\newcommand{\dpa}{\ensuremath{\mathtt{DP}}}

\newcommand{\mdpa}[1]{\ensuremath{\mathtt{PCNT}_{#1}}}

\newcommand{\eqdef}{\ensuremath{\,\mathrel{\mathop:}=}}

\newcommand{\Card}[1]{|#1|}
\renewcommand{\P}{\text{\normalfont P}\xspace}

\newcommand{\PRIM}{\AlgS} 

\newcommand{\SAT}{\textsc{Sat}\xspace}

\newcommand{\QBFSAT}{\textsc{QSat}\xspace}

\newcommand{\Q}{\ensuremath{Q}}

\newcommand{\INC}{\ensuremath{{\algo{INC}}}\xspace}

\newcommand{\problemFont}[1]{\textsc{#1}}

\newlength\problemlength
\newcommand\dproblem[3]{%
\begin{center}
\fbox{%
\begin{minipage}{.93	\linewidth}%
\begin{list}{}{\labelwidth\problemlength \labelsep.7em \rightmargin1.5em
\leftmargin\problemlength \advance\leftmargin by3em
\parsep0ex \itemsep.2ex plus.1ex}
\item[{\sl Problem:\hfill}] {\problemFont{#1}}
\item[{\sl Input:  \hfill}] #2
\item[{\sl Task: \hfill}] #3
\end{list}
\end{minipage}
}
\end{center}
}

\newcommand{\AlgS}{\algo{SAT}\xspace}%
\begin{document}

\maketitle

\begin{abstract}%
  In this paper, we introduce a novel algorithm to solve
  \emph{projected model counting} (\PMC). \PMC asks to count solutions
  of a Boolean formula with respect to a given set of \emph{projected
    variables}, where multiple solutions that are identical when
  restricted to the projected variables count as only one solution.
  Our algorithm exploits small treewidth of the primal 
  graph of the input instance. It runs in time~$\bigO{2^{2^{k+4}} n^2}$
  where $k$ is the treewidth and $n$ is the input size of the
  instance. In other words, we obtain that the problem~\PMC is
  fixed-parameter tractable when parameterized by treewidth.  Further,
  we take the exponential time hypothesis (ETH) into consideration and
  establish lower bounds of bounded treewidth algorithms for \PMC,
  yielding asymptotically tight runtime bounds of our algorithm.


  \springerversion{}
\end{abstract}

\section{Introduction}\label{sec:introduction}
A problem that has been used to solve a large variety of real-world
questions is the \emph{model counting problem}
(\cSAT)~\cite{AbramsonBrownEdwards96a,ChoiBroeckDarwiche15a,DomshlakHoffmann07a,MeelEtAl17a,ManningRaghavanSchutze08a,PourretNaimBruce08a,SahamiDumaisHeckerman98a,SangBeameKautz05a,XueChoiDarwiche12a}.
It asks to compute the number of solutions of a Boolean
formula~\cite{GomesKautzSabharwalSelman08a} and is theoretically of
high worst-case complexity
($\cntc\P$-complete~\cite{Valiant79,Roth96a}). Lately, both \cSAT and
its approximate version have received renewed attention in theory and
practice~\cite{ChakrabortyMeelVardi16a,MeelEtAl17a,LagniezMarquis17a,SaetherTelleVatshelle15a}.
A concept that allows very natural abstractions of data and query
results is projection. Projection has wide applications in
databases~\cite{AbiteboulHullVianu95} and declarative problem
modeling.  
The problem \emph{projected model counting} (\PMC) asks to count
solutions of a Boolean formula with respect to a given set of
\emph{projected variables}, 
where multiple solutions that are identical when restricted to the
projected variables count as only one solution.
If all variables of the formula are projected variables, then \PMC is
the \cSAT problem and if there are no projected variables then it is
simply the \SAT problem.
Projected variables allow for solving problems where one needs to
introduce auxiliary variables, in particular, if these variables are
functionally independent of the variables of interest, in the problem
encoding,~e.g.,~\cite{GebserSchaubThieleVeber11,GinsbergParkesRoy98a}.
%
%

When we consider the computational complexity of \PMC it turns out
that under standard assumptions the problem is even harder than \cSAT,
more precisely, complete for the class
$\cntc\NP$~\cite{DurandHermannKolaitis05}.
Even though there is a \PMC solver~\cite{AzizChuMuise15a} and an \ASP
solver that implements projected
enumeration~\cite{GebserKaufmannSchaub09a}, \PMC has received very
little attention in parameterized algorithmics so far.
Parameterized
algorithms~\cite{CyganEtAl15,DowneyFellows13,FlumGrohe06,Niedermeier06}
tackle computationally hard problems by directly exploiting certain
structural properties (parameter) of the input instance to solve the
problem faster, preferably in polynomial-time for a fixed parameter
value.
%
%
%
In this paper, we consider the treewidth of graphs associated with the
given input formula as parameter, namely the primal 
graph~\cite{SamerSzeider10b}.
Roughly speaking, small \emph{treewidth} of a graph measures its
tree-likeness and sparsity. Treewidth is defined in terms of
\emph{tree decompositions (TDs)}, which are arrangements of graphs
into trees.
%
When we take advantage of small treewidth, we usually take a TD and
evaluate the considered problem in parts, via \emph{dynamic
  programming~(DP)} on the TD. 
%

\paragraph{New Contributions.}
%
\begin{enumerate}
\item We introduce a novel algorithm to \emph{solve projected model
    counting (\PMC)} in time~$\bigO{2^{2^{k+4}} n^2}$ where $k$ is the
  treewidth of the primal 
  graph of the instance and $n$ is the size of the input instance.
  Similar to recent DP algorithms for problems on the second level of
  the polynomial hierarchy~\cite{FichteEtAl17b}, our algorithm
  traverses the given tree decomposition multiple times (multi-pass).
  In the first traversal, we run a dynamic programming algorithm on
  tree decompositions to solve \SAT~\cite{SamerSzeider10b}. In a
  second traversal, we construct equivalence classes on top of the
  previous computation to obtain model counts with respect to the
  projected variables by exploiting combinatorial properties of
  intersections.
\item We establish that our \emph{runtime bounds are asymptotically tight under the
    exponential time hypothesis (ETH)}~\cite{ImpagliazzoPaturiZane01}
  using a recent result by Lampis and Mitsou~\cite{LampisMitsou17},
  who established lower bounds for the
  problem~$\exists\forall$\hy\SAT assuming ETH.
  Intuitively, ETH states a complexity theoretical lower bound on how
  fast satisfiability problems can be solved. More precisely, one
  \emph{cannot} solve 3\hy\SAT in
  time~$2^{s\cdot n}\cdot n^{\bigO{1}}$ for some~$s>0$ and number~$n$
  of variables.
\end{enumerate}
%
%

%
%

\section{Preliminaries}\label{sec:preliminaries}


For a set~$X$, let $\ta{X}$ be the \emph{power set of~$X$}
consisting of all subsets~$Y$ with $\emptyset \subseteq Y \subseteq X$.
%
%
Recall the well-known combinatorial inclusion-exclusion
principle~\cite{GrahamGrotschelLovasz95a}, which states that for two
finite sets~$A$ and $B$ it is true
that~$\Card{A \cup B} = \Card{A} + \Card{B} - \Card{A \cap B}$. Later,
we need a generalized version for arbitrary many sets. Given for some
integer~$n$ a family of finite sets~$X_1$, $X_2$, $\ldots$,
$X_n$, 
the number of elements in the union
over all sets is
$\Card{\bigcup^n_{j = 1} X_j} = \sum_{I \subseteq \{1, \ldots, n\}, I
  \neq \emptyset} (-1)^{\Card{I}-1} \Card{\bigcap_{i \in I}
  X_i}$. 

\paragraph{Satisfiability.}
A literal is a (Boolean) variable~$x$ or its negation~$\neg x$. A
\emph{clause} is a finite set of literals, interpreted as the
disjunction of these literals.
%
%
A \emph{(CNF) formula} is a finite set of clauses, interpreted as the
conjunction of its clauses.  A 3\hy CNF has clauses of length at
most~3. 
Let $F$ be a formula.  A \emph{sub-formula~$S$} of~$F$ is a
subset~$S\subseteq F$ of~$F$.  For a clause~$c \in F$, we let
$\var(c)$ consist of all variables that occur in~$c$ and
$\var(F)\eqdef\bigcup_{c \in F} \var(c)$.  A (partial)
\emph{assignment} is a mapping $\alpha: \var(F) \rightarrow \{0,1\}$.
For $x\in \var(F),$ we define $\alpha(\neg x) \eqdef 1 - \alpha(x)$.
The formula~$F$ \emph{under the assignment~$\alpha \in \ta{\var(F)}$} 
is the formula~$F_{|\alpha}$ obtained from~$F$ by removing all
clauses~$c$ containing a literal set to~$1$ by $\alpha$ and removing
from the remaining clauses all literals set to~$0$ by $\alpha$. An
assignment~$\alpha$ is \emph{satisfying} if $F_{|\alpha}=\emptyset$ and
$F$ is \emph{satisfiable} if there is a satisfying
assignment~$\alpha$. %
Let $V$ be a set of variables. An \emph{interpretation} is a
set~$J\subseteq V$ and its induced assignment~$\alpha_{J,V}$ of~$J$
with respect to $V$ is defined as
follows~$\alpha_{J,V} \eqdef \SB v \mapsto 1 \SM v \in J \cap V \SE
\cup \SB v \mapsto 0 \SM v \in V \setminus J \SE$.
We simply write $\alpha_{J}$ for $\alpha_{J,V}$ if $V=\var(F)$.
An interpretation~$J$ is a \emph{model} of~$F$, denoted by~$J\models F$, if its
induced assignment~$\alpha_J$ is satisfying.
%
%
%
%
Given a formula~$F$; the problem \SAT asks whether $F$ is satisfiable
and the problem \cSAT asks to output the number of models
of~$F$,~i.e., $\Card{S}$ where $S$ is the set of all models of~$F$.


\paragraph{Projected Model Counting.} %
An instance of the projected model counting problem is a pair~$(F,P)$
where $F$ is a (CNF) formula and $P$ is a set of Boolean variables
such that $P \subseteq\var(F)$.  We call the set~$P$ \emph{projection
  variables} of the instance. The \emph{projected model count} of a
formula~$F$ with respect to~$P$ is the number of total
assignments~$\alpha$ to variables in~$P$ such that the
formula~$F_{|\alpha}$ under~$\alpha$ is satisfiable.
The \emph{projected model counting problem
  (\PMC)}~\cite{AzizChuMuise15a} asks to output the projected model
count of~$F$,~i.e., $\Card{ \SB M \cap P \SM M \in S \SE}$ where $S$
is the set of all models of~$F$.

\begin{example}\label{ex:running0}
  Consider
  formula~$F\eqdef \{\overbrace{\neg a \vee b \vee p_1}^{c_1},
  \overbrace{a\vee \neg b \vee \neg p_1}^{c_2}, \overbrace{a \vee
    p_2}^{c_3}, \overbrace{a \vee \neg p_2}^{c_4}\}$ and
  set~$P\eqdef\{p_1,p_2\}$ of projection variables.
  The models of formula~$F$ are $\{a,b\}$, $\{a,p_1\}$,
  $\{a,b,p_1\}$,$\{a,b,p_2\}$, $\{a,p_1,p_2\}$, and $\{a,b,p_1,p_2\}$.
  However, projected to the set~$P$, we only have models $\emptyset$,
  $\{p_1\}$, $\{p_2\}$, and $\{p_1,p_2\}$.
  Hence, the model count of~$F$ is 6 whereas the projected model count
  of instance~$(F,P)$ is 4.
\end{example}



%




\newcommand{\restrict}[2]{\ensuremath{#1\cap #2}}


\paragraph{Computational Complexity.}
We assume familiarity with standard notions in computational
complexity~\cite{Papadimitriou94}
and use counting complexity classes as defined by Hesaspaandra and
Vollmer~\cite{HemaspaandraVollmer95a}.
%
%
%
For parameterized complexity, we refer to standard
texts~\cite{CyganEtAl15,DowneyFellows13,FlumGrohe06,Niedermeier06}.
%
%
Let $\Sigma$ and $\Sigma'$ be some finite alphabets.  We call
$I \in \Sigma^*$ an \emph{instance} and $\CCard{I}$ denotes the size
of~$I$.  
%
Let $L \subseteq \Sigma^* \times \Nat$ and
$L' \subseteq {\Sigma'}^*\times \Nat$ be two parameterized problems. An
\emph{fpt-reduction} $r$ from $L$ to $L'$ is a many-to-one reduction
from $\Sigma^*\times \Nat$ to ${\Sigma'}^*\times \Nat$ such that for all
$I \in \Sigma^*$ we have $(I,k) \in L$ if and only if
$r(I,k)=(I',k')\in L'$ such that $k' \leq g(k)$ for a fixed computable
function $g: \Nat \rightarrow \Nat$, and there is a computable function
$f$ and a constant $c$ such that $r$ is computable in time
$O(f(k)\CCard{I}^c)$~\cite{FlumGrohe06}.
A \emph{witness function} is a
function~$\mathcal{W}\colon \Sigma^* \rightarrow 2^{{\Sigma'}^*}$ that
maps an instance~$I \in \Sigma^*$ to a finite subset
of~${\Sigma'}^*$. We call the set~$\WWW(I)$ the \emph{witnesses}. A
\emph{parameterized counting
  problem}~$L: \Sigma^* \times \NAT_0 \rightarrow \Nat_0$ is a
function that maps a given instance~$I \in \Sigma^*$ and an
integer~$k \in \NAT$ to the cardinality of its
witnesses~$\card{\WWW(I)}$.
We call $k$ the \emph{parameter}.
%
%
%
The \emph{exponential time hypothesis} (ETH) states
that 
the (decision) problem~\SAT on 3\hy CNF formulas \emph{cannot} be
solved in time $2^{s\cdot n}\cdot n^{\bigO{1}}$ for some~$s>0$ where
$n$ is the number of variables~\cite{ImpagliazzoPaturiZane01}.
%

\paragraph{Tree Decompositions and Treewidth.} %
For basic terminology on graphs
, we refer to standard
texts~\cite{Diestel12,BondyMurty08}.  For a tree~$T=(N,A,n)$ with
root~$n$ and a node~$t \in N$, we let $\children(t, T)$ be the
sequence of all nodes~$t'$ in arbitrarily but fixed order, which have
an edge~$(t,t') \in A$.
Let $G=(V,E)$ be a graph.
A \emph{tree decomposition (TD)} of graph~$G$ is a pair
$\TTT=(T,\chi)$ where $T=(N,A,n)$ is a rooted tree, $n\in N$ the root,
and $\chi$ a mapping that assigns to each node $t\in N$ a set
$\chi(t)\subseteq V$, called a \emph{bag}, such that the following
conditions hold:
(i)~$V=\bigcup_{t\in N}\chi(t)$ and
$E \subseteq\bigcup_{t\in N}\SB uv \SM u,v\in \chi(t)\SE$; 
(ii)
for each $r, s, t\in N$ such that $s$ lies on the path from $r$ to
$t$, we have $\chi(r) \cap \chi(t) \subseteq \chi(s)$.
Then, $\width(\TTT) \eqdef \max_{t\in N}\Card{\chi(t)}-1$.  The
\emph{treewidth} $\tw{G}$ of $G$ is the minimum $\width({\TTT})$ over
all tree decompositions $\TTT$ of $G$.
For arbitrary but fixed $w \geq 1$, it is feasible in linear time to
decide if a graph has treewidth at most~$w$ and, if so, to compute a
tree decomposition of width $w$~\cite{Bodlaender96}.
In order to simplify case distinctions in the algorithms, we always
use so-called nice tree decompositions, which can be computed in
linear time without increasing the width~\cite{BodlaenderKoster08} and
are defined as follows.
For a node~$t \in N$, we say that $\type(t)$ is $\leaf$ if
$\children(t,T)=\langle \rangle$; $\join$ if
$\children(t,T) = \langle t',t''\rangle$ where
$\chi(t) = \chi(t') = \chi(t'') \neq \emptyset$; $\intr$
(``introduce'') if $\children(t,T) = \langle t'\rangle$,
$\chi(t') \subseteq \chi(t)$ and $|\chi(t)| = |\chi(t')| + 1$; $\rem$
(``removal'') if $\children(t,T) = \langle t'\rangle$,
$\chi(t') \supseteq \chi(t)$ and $|\chi(t')| = |\chi(t)| + 1$. If for
every node $t\in N$, $\type(t) \in \{ \leaf, \join, \intr, \rem\}$ and
bags of leaf nodes and the root are empty, then the TD is called
\emph{nice}.

\section{Dynamic Programming on TDs for SAT}

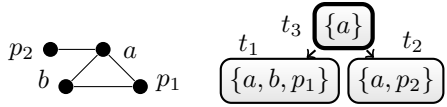
\begin{figure}[t]%
\centering
\begin{tikzpicture}[node distance=7mm,every node/.style={fill,circle,inner sep=2pt}]
\node (a) [label={[text height=1.5ex,yshift=0.0cm,xshift=0.05cm]left:$p_2$}] {};
\node (b) [right of=a,label={[text height=1.5ex]right:$a$}] {};
\node (c) [below left of=b,label={[text height=1.5ex,yshift=0.09cm,xshift=0.05cm]left:$b$}] {};
\node (d) [below right of=b,label={[text height=1.5ex,yshift=0.09cm,xshift=-0.05cm]right:$p_1$}] {};
\draw (a) to (b);
\draw (b) to (c);
\draw (b) to (d);
\draw (c) to (d);
\end{tikzpicture}\hspace{1em}%
\begin{tikzpicture}[node distance=0.5mm]
\tikzset{every path/.style=thick}

\node (leaf1) [tdnode,label={[yshift=-0.25em,xshift=0.5em]above left:$t_1$}] {$\{a,b,p_1\}$};
\node (leaf2) [tdnode,label={[xshift=-1.0em, yshift=-0.15em]above right:$t_2$}, right = 0.1cm of leaf1]  {$\{a,p_2\}$};
\coordinate (middle) at ($ (leaf1.north east)!.5!(leaf2.north west) $);
\node (join) [tdnode,ultra thick,label={[]left:$t_3$}, above  = 1mm of middle] {$\{a\}$};

\coordinate (top) at ($ (join.north east)+(3.5em,0) $);
\coordinate (bot) at ($ (top)+(0,-4em) $);

\draw [->] (join) to (leaf1);
\draw [->] (join) to (leaf2);
\end{tikzpicture}%
\caption{Primal graph~$P_F$ of~$F$ from Example~\ref{ex:running1}
  (left) with a TD~${\cal T}$ of graph~$P_F$
  (right).}%
\label{fig:graph-td}%
\end{figure}

Before we introduce our algorithm, we need some notations for dynamic
programming on tree decompositions and recall how to solve the
decision problem~\SAT by exploiting small treewidth.


%
%
%
%
%
%
%
%

\paragraph{Graph Representation of \SAT Formulas.}
In order to use tree decompositions for satisfiability problems, we
need a dedicated graph representation of the given formula~$F$.
The \emph{primal graph}~$P_F$ of~$F$ has as vertices the variables of~$F$
and two variables are joined by an edge if they occur together in a
clause of~$F$.
%
%
%
%
Further, we define some auxiliary notation.  For a given node~$t$ of a
tree decomposition~$(T,\chi)$ of the primal graph, we let
$F_t \eqdef \SB c \SM c \in F, \var(c) \subseteq \chi(t)\SE$,~i.e.,
clauses entirely covered by~$\chi(t)$.  The set~$\progt{t}$ denotes
the union over~$F_{s}$ for all descendant nodes~$s\in N$ of~$t$.
%
%
%
In the following, we sometimes simply write \emph{tree decomposition of
  formula~$F$} or \emph{treewidth of~$F$} and omit the actual graph
representation of~$F$.

\begin{example}\label{ex:running1}
  Consider formula~$F$ from Example~\ref{ex:running0}.
  The primal graph~$P_F$ of formula~$F$ and a tree
  decomposition~$\TTT$ of~$P_F$ are depicted in
  Figure~\ref{fig:graph-td}. Intuitively, ${\cal T}$ allows to
  evaluate formula~$F$ in parts. When evaluating $F_{\leq t_3}$, we
  split into $F_{\leq t_1}=\{c_1,c_2\}$ and
  $F_{\leq t_2}=\{c_3, c_4\}$, respectively.
\end{example}
%
%
%

%
%

%
%
  

%



\longversion{
\begin{figure}[t]
\centering
\includegraphics[scale=0.8]{figure.pdf}
\caption{The DP approach, where table algorithm~$\algo{A}$ modifies
  tables.~\cite{FichteEtAl17a}}
\label{fig:framework}
\end{figure}%
}

\paragraph*{Dynamic Programming on TDs.} %
Algorithms that solve \SAT or \cSAT~\cite{SamerSzeider10b} in linear
time for input formulas of bounded treewidth proceed by dynamic
programming along the tree decomposition (in post-order) where at each
node~$t$ of the tree information is gathered~\cite{BodlaenderKloks96}
in a table~$\tau_t$.
A \emph{table}~$\tab{}$ is a set of rows, where a
\emph{row}~$\vec\tabval \in \tab{}$ is a sequence of fixed length. 
Tables are derived by an algorithm, which we therefore call
\emph{table algorithm}~$\AlgA$.  
The actual length, content, and meaning of the rows depend on the
algorithm~$\AlgA$ that derives tables.  Therefore, we often explicitly
state \emph{$\AlgA$-row} if rows of this \emph{type} are syntactically
used for table algorithm~$\AlgA$ and similar \emph{$\AlgA$-table} for
tables.
For sake of comprehension, we
specify the rows before presenting the actual table algorithm for
manipulating tables.
The rows used by a table algorithm~$\AlgS$ have in
common that the first position of these rows manipulated by~$\AlgS$
consists of an interpretation. The remaining positions of the row
depend on the considered table algorithm.
For each sequence~$\vec \tabval \in \tab{}$, we
write~$I(\vec \tabval)$ to address the interpretation (first) part of
the sequence~$\vec\tabval$. Further, for a given positive integer~$i$,
we denote by $\vec\tabval_{(i)}$ the $i$-th element of
row~$\vec\tabval$ and define~$\tab{(i)}$ as
$\tab{(i)}\eqdef\{\vec u_{(i)} \mid \vec u \in \tab{}\}$.

%
Then, the dynamic programming approach for propositional
satisfiability \longversion{works as outlined in Figure~\ref{fig:framework} and}
performs the following steps:
\begin{enumerate}
\item Construct the primal graph 
$P_F$ of~$F$. 
\item Compute a tree decomposition~$(T,\chi)$ of~$P_F$,
  where~$T=(N,\cdot,n)$.
\item\label{step:dp} Run $\dpa_\AlgS$ (see Listing~\ref{fig:dpontd}),
  which executes a table algorithm~$\AlgS$ for every node~$t$ in post-order of the nodes
  in~$N$, and returns $\ATab{\AlgS}$ mapping every node~$t$ to its table. $\AlgS$ takes as
  input\footnote{Actually, \AlgS takes in addition as input \Prev,
    which contains a mapping of nodes of the tree decomposition to
    tables,~i.e., tables of the previous pass. Later, we use this for
    a second traversal to pass results ($\ATab{\AlgS}$) from the first
    traversal to the table algorithm~$\PROJ$ for projected model
    counting in the second traversal.
  } %
  bag~$\chi(t)$, sub-formula~$F_t$, and tables \Tab{} previously
  computed at children of~$t$ and outputs a
  table~$\tab{t}$. 
\item Print the result by interpreting the table for root~$n$ of~$T$.
\end{enumerate}

\noindent%
Listing~\ref{fig:prim} presents table algorithm~\PRIM that uses
the primal graph representation. We provide only brief intuition, for
details we refer to the original source~\cite{SamerSzeider10b}.
The main idea is to store in table~$\tab{t}$ only interpretations that
are a model of sub-formula~$F_{\leq t}$ when restricted to
bag~$\chi(t)$.
Table algorithm~\PRIM transforms at node~$t$ certain row
combinations of the tables ($\Tab{}$) of child nodes of~$t$ into rows
of table~$\tab{t}$. The transformation depends on a case where
variable~$a$ is added or not added to an interpretation ($\intr$),
removed from an interpretation ($\rem$), or where coinciding
interpretations are required ($\join$).  In the end, an
interpretation~$I(\vec u)$ from a row~$\vec u$ of the table~$\tau_n$
at the root~$n$ proves that there is a supset~$J \supseteq I(\vec u)$
that is a model of~$F = \progt{n}$, and hence that the formula is
satisfiable.%
%
%
%
%
\begin{algorithm}[t]%
  \KwData{%
    Table algorithm $\AlgA$, TD~$\TTT=(T,\chi)$ of~$F$ s.t.
    $T=(N,\cdot,n)$, tables~\Prev.\hspace{-5em}
  }%
  \KwResult{%
    Table $\ATab{\AlgA}$, which maps each TD node~$t\in N$ to some computed
    table~$\tau_t$.\hspace{-5em}
  } %
  \For{\text{\normalfont iterate} $t$ in \text{\normalfont post-order}(T,n)}{
    \vspace{-0.05em}%

    $\Tab{} \eqdef \langle \ATab{\AlgA}[t_1],\ldots,
    \ATab{\AlgA}[t_\ell] \rangle$ where
    $\children(t, T) = \langle t_1, \ldots, t_\ell
    \rangle\hspace{-5em}$

    $\ATab{\AlgA}[t] \leftarrow {\AlgA}(t,\chi(t),\prog_t,P \cap
    \chi(t), \Tab{},\Prev)$ %
    \vspace{-0.5em} %
  }%
  \Return{$\ATab{\AlgA}$} \vspace{-0.2em}%
 \caption{Algorithm ${\dpa}_{\AlgA}( (F,P), \TTT, \Prev)$ for DP on
   TD~${\cal T}$~\protect\cite{FichteEtAl17a}.}
\label{fig:dpontd}
\end{algorithm}%
\renewcommand{\eqdef}{\leftarrow}
%
%
 \begin{algorithm}[t]
   \KwData{Node~$t$, bag $\chi_t$, clauses~$\prog_t$, sequence~$\Tab{}$ of tables.{~\bf Out:} Table~$\tab{t}.\hspace{-5em}$}
   \lIf(\hspace{-1em})
   {$\type(t) = \leaf$}{%
     $\tab{t} \eqdef \{ \langle
     \tuplecolor{\inputPredColor}{\emptyset}
     \rangle \}$%
     %
   }%
  \uElseIf{$\type(t) = \intr$, $a\hspace{-0.1em}\in\hspace{-0.1em}\chi_t$ is introduced, and $\Tab{} = \langle \tau' \rangle$}{
   \vspace{-0.05em}
   \makebox[3.19cm][l]{$\tab{t} \eqdef \{ \langle \tuplecolor{\inputPredColor}{K} \rangle$}
     $|\;\langle \tuplecolor{\inputPredColor}{J} \rangle \in \tab{}',  {{\tuplecolor{\inputPredColor}{K \in \{J, J \cup \{a\}\}}, \tuplecolor{\inputPredColor}{K}}} \models \prog_t
      \} \hspace{-5em}$
     %
   \vspace{-0.05em}
     }\vspace{-0.05em}%
     \uElseIf{$\type(t) = \rem$, $a \not\in \chi_t$ is removed, and $\Tab{} = \langle \tau'\rangle$}{%
       \makebox[3.3cm][l]{$\tab{t} \eqdef \{ \langle \tuplecolor{\inputPredColor}{J \setminus \{a\}}
       \rangle$}$|\;\langle \tuplecolor{\inputPredColor}{J}
       \rangle \in \tab{}' \}\hspace{-5em}$
       \vspace{-0.1em}
     } %
     \uElseIf{$\type(t) = \join$, and $\Tab{} = \langle \tau', \tau'' \rangle$}{%
       \makebox[3.3cm][l]{$\tab{t} \eqdef \{ \langle \tuplecolor{\inputPredColor}{J}
         \rangle$}$|\;\langle \tuplecolor{\inputPredColor}{J} \rangle \in \tab{}', \langle \tuplecolor{\inputPredColor}{J} \rangle \in \tab{}''
       \}\hspace{-5em}$
       \vspace{-0.1em}
     } 
     \Return $\tab{t}$
     \vspace{-0.25em}
     \caption{Table
       algorithm~$\algo{SAT}(t, \chi_t,\prog_t,\cdot, \Tab{},
       \cdot)$~\protect\cite{SamerSzeider10b}.}
 \label{fig:prim}
\end{algorithm}%
\renewcommand{\eqdef}{{\ensuremath{\,\mathrel{\mathop:}=}}}
%
%
%
%
%
Example~\ref{ex:sat} lists selected tables when running
algorithm~$\dpa_{\PRIM}$.
\begin{example}\label{ex:sat}
  Consider formula~$\prog$ from Example~\ref{ex:running1}.
  Figure~\ref{fig:running1} illustrates a tree
  decomposition~$\TTT'=(\cdot, \chi)$ of the primal graph of~$F$ and
  tables~$\tab{1}$, $\ldots$, $\tab{12}$ that are obtained during the
  execution of~$\dpa_{\PRIM}((F,\cdot),\TTT',\cdot)$.
  We assume that each row in a table $\tab{t}$ is identified by a
  number,~i.e., row $i$ corresponds to
  $\vec{u_{t.i}} = \langle J_{t.i} \rangle$.

  Table~$\tab{1}=\SB \langle\emptyset\rangle \SE$ as
  $\type(t_1) = \leaf$.
  Since $\type(t_2) = \intr$, we construct table~$\tab{2}$
  from~$\tab{1}$ by taking~$J_{1.i}$ and $J_{1.i}\cup \{a\}$ for
  each~$\langle J_{1.i}\rangle \in \tab{1}$. Then,
  $t_3$ introduces $p_1$ and $t_4$ introduces $b$.
  $\prog_{t_1}=\prog_{t_2}=\prog_{t_3} = \emptyset$, but since
  $\chi(t_4) \subseteq \var(c_1)$ we have
  $\prog_{t_4} = \{c_1,c_2\}$ for $t_4$.
  In consequence, for each~$J_{4.i}$ of table~$\tab{4}$, we have
  $\{c_1,c_2\} \models {{J_{4.i}}}$ since \PRIM enforces
  satisfiability of $\prog_t$ in node~$t$.  
  Since $\type(t_5) = \rem$, we remove variable~$p_1$ from all
  elements in $\tab{4}$ to construct $\tab{5}$. Note that we have
  already seen all rules where $p_1$ occurs and hence $p_1$ can no
  longer affect interpretations during the remaining traversal. We
  similarly create $\tab{6}=\{\langle \emptyset \rangle, \langle a \rangle\}$
  and~$\tab{{10}}=\{\langle a \rangle\}$.
  Since $\type(t_{11})=\join$, we build table~$\tab{11}$ by taking
  the intersection of $\tab{6}$ and $\tab{{10}}$. Intuitively, this
  combines interpretations agreeing on~$a$.
  %
  %
  By definition (primal graph and TDs), for every~$c \in \prog$,
  variables~$\var(c)$ occur together in at least one common bag.
  Hence, $\prog=\progt{t_{12}}$ and since
  $\tab{12} = \{\langle \emptyset \rangle \}$, we can reconstruct for example
  model~$\{a,b,p_2\} = J_{11.1} \cup J_{5.4} \cup J_{9.2}$ of~$F$ using highlighted (yellow) rows in Figure~\ref{fig:running1}.
  On the other hand, if~$F$ was unsatisfiable, $\tab{12}$ would be empty ($\emptyset$). 
  %
%
\end{example}%

\begin{figure}[t]
\centering
\begin{tikzpicture}[node distance=0.5mm]
\tikzset{every path/.style=thick}

\node (l1) [stdnode,label={[tdlabel, xshift=0em,yshift=+0em]right:${t_1}$}]{$\emptyset$};
\node (i1) [stdnode, above=of l1, label={[tdlabel, xshift=0em,yshift=+0em]right:${t_2}$}]{$\{a\}$};
\node (i12) [stdnode, above=of i1, label={[tdlabel, xshift=0em,yshift=+0em]right:${t_3}$}]{$\{a,p_1\}$};
\node (i13) [stdnode, above=of i12, label={[tdlabel, xshift=0em,yshift=+0em]right:${t_4}$}]{$\{a,b,p_1\}$};
\node (r1) [stdnode, above=of i13, label={[tdlabel, xshift=0em,yshift=+0em]right:${t_5}$}]{$\{a,b\}$};
\node (r12) [stdnode, above=of r1, label={[tdlabel, xshift=0em,yshift=+0em]right:${t_6}$}]{$\{a\}$};
\node (l2) [stdnode, right=2.5em of i12, label={[tdlabel, xshift=0em,yshift=+0em]left:${t_7}$}]{$\emptyset$};
\node (i2) [stdnode, above=of l2, label={[tdlabel, xshift=0em,yshift=+0em]left:${t_8}$}]{$\{p_2\}$};
\node (i22) [stdnode, above=of i2, label={[tdlabel, xshift=0em,yshift=+0em]left:${t_9}$}]{$\{a,p_2\}$};
\node (r2) [stdnode, above=of i22, label={[tdlabel, xshift=0em,yshift=+0em]left:${t_{10}}$}]{$\{a\}$};
\node (j) [stdnode, above left=of r2, yshift=-0.25em, label={[tdlabel, xshift=0em,yshift=+0.15em]right:${t_{11}}$}]{$\{a\}$};
\node (rt) [stdnode,ultra thick, above=of j, label={[tdlabel, xshift=0em,yshift=+0em]right:${t_{12}}$}]{$\emptyset$};
\node (label) [font=\scriptsize,left=of rt]{${\cal T}'$:};
\node (leaf1) [stdnode, left=1.25em of i1, yshift=0.5em, label={[tdlabel, xshift=2.75em,yshift=+1.25em]above left:$\tab{4}$}]{%
	\begin{tabular}{l}%
		\multicolumn{1}{l}{$\langle \tuplecolor{\inputPredColor}{J_{4.i}} \rangle$}\\
		\hline\hline
		$\langle \tuplecolor{\inputPredColor}{\emptyset}\rangle$\\\hline
		$\langle \tuplecolor{\inputPredColor}{\{b\}}\rangle$\\\hline
		\rowcolor{yellow}$\langle\tuplecolor{\inputPredColor}{\{a,b\}}\rangle$\\\hline
		$\langle \tuplecolor{\inputPredColor}{\{p_1\}}\rangle$\\\hline
		$\langle\tuplecolor{\inputPredColor}{\{a,p_1\}}\rangle$\\\hline
		$\langle\tuplecolor{\inputPredColor}{\{a,b,p_1\}}\rangle$\\
	\end{tabular}%
};
\node (leaf1b) [stdnodenum,left=of leaf1,xshift=0.6em,yshift=0pt]{%
	\begin{tabular}{c}%
		\multirow{1}{*}{$i$}\\ 
		\hline\hline
		$1$ \\\hline
		$2$ \\\hline
		$3$ \\\hline
		$4$ \\\hline
		$5$ \\\hline
		$6$
	\end{tabular}%
};
\node (leaf0x) [stdnode, left=0.75em of leaf1b, yshift=1.5em, label={[tdlabel, xshift=2em,yshift=+1em]above left:$\tab{5}$}]{%
	\begin{tabular}{l}%
		\multicolumn{1}{l}{$\langle \tuplecolor{\inputPredColor}{J_{5.i}} \rangle$}\\
		\hline\hline
		$\langle \tuplecolor{\inputPredColor}{\emptyset}\rangle$\\\hline
		$\langle\tuplecolor{\inputPredColor}{\{a\}}\rangle$\\\hline
		$\langle\tuplecolor{\inputPredColor}{\{b\}}\rangle$\\\hline
		\rowcolor{yellow}$\langle\tuplecolor{\inputPredColor}{\{a,b\}}\rangle$\\
	\end{tabular}%
};
\node (leaf0b) [stdnodenum,left=of leaf0x,xshift=0.6em,yshift=0pt]{%
	\begin{tabular}{c}%
		\multirow{1}{*}{$i$}\\ 
		\hline\hline
		$1$ \\\hline
		$2$ \\\hline
		$3$ \\\hline
		$4$ 
	\end{tabular}%
};
\node (leaf2b) [stdnodenum,right=2.5em of j,xshift=-0.75em,yshift=+0.25em]  {%
	\begin{tabular}{c}%
		\multirow{1}{*}{$i$}\\ 
		\hline\hline
		$1$\\\hline
		$2$\\
	\end{tabular}%
};
\node (leaf2) [stdnode,right=-0.4em of leaf2b, label={[tdlabel, xshift=0em,yshift=-0.25em]below:$\tab{9}$}]  {%
	\begin{tabular}{l}%
		\multirow{1}{*}{$\langle \tuplecolor{\inputPredColor}{J_{9.i}} \rangle$}\\ 
		\hline\hline
		$\langle \tuplecolor{\inputPredColor}{\{a\}}\rangle$\\\hline
		\rowcolor{yellow}$\langle \tuplecolor{\inputPredColor}{\{a,p_2\}}\rangle$\\
	\end{tabular}%
};
\coordinate (middle) at ($ (leaf1.north east)!.5!(leaf2.north west) $);
\node (join) [stdnode,left=4.5em of r12, yshift=0.5em, label={[tdlabel, xshift=2em,yshift=+0.25em]above left:$\tab{{11}}$}] {%
	\begin{tabular}{l}%
		\multirow{1}{*}{$\langle \tuplecolor{\inputPredColor}{J_{11.i}} \rangle$}\\
		\hline\hline
		\rowcolor{yellow}$\langle \tuplecolor{\inputPredColor}{\{a\}} \rangle$\\
	\end{tabular}
};
\node (joinb) [stdnodenum,left=-0.45em of join] {%
	\begin{tabular}{c}
		\multirow{1}{*}{$i$}\\
		\hline\hline
		$1$\\
	\end{tabular}%
};
\node (rtx) [stdnode,left=0.0em of r12, yshift=2.75em, label={[tdlabel, xshift=0em,yshift=-0.8em]right:$\tab{{12}}$}] {%
	\begin{tabular}{l}%
		\multirow{1}{*}{$\langle \tuplecolor{\inputPredColor}{J_{12.i}} \rangle$}\\
		\hline\hline
		\rowcolor{yellow}$\langle \tuplecolor{\inputPredColor}{\emptyset} \rangle$\\
	\end{tabular}
};
\node (rtb) [stdnodenum,left=-0.45em of rtx] {%
	\begin{tabular}{c}
		\multirow{1}{*}{$i$}\\
		\hline\hline
		$1$\\
	\end{tabular}%
};
\node (leaf0n) [stdnodenum,yshift=0.5em, right=2.5em of l1] {%
	\begin{tabular}{c}%
		\multirow{1}{*}{$i$}\\ 
		\hline\hline
		$1$
	\end{tabular}%
};
\node (leaf0) [stdnode,right=-0.5em of leaf0n, label={[tdlabel, xshift=-1em,yshift=0.15em]above right:$\tab{1}$}] {%
	\begin{tabular}{l}%
		\multicolumn{1}{l}{$\langle \tuplecolor{\inputPredColor}{M_{1.i}} \rangle$}\\
		\hline\hline
		\rowcolor{yellow}$\langle \tuplecolor{\inputPredColor}{\emptyset}\rangle$
	\end{tabular}%
};
\coordinate (top) at ($ (leaf2.north east)+(0.6em,-0.5em) $);
\coordinate (bot) at ($ (top)+(0,-12.9em) $);

\draw [<-] (j) to (rt);
\draw [->] (j) to ($ (r12.north)$);
\draw [->] (j) to ($ (r2.north)$);
\draw [->](r2) to (i22);
\draw [<-](i2) to (i22);
\draw [<-](l2) to (i2);
\draw [<-](l1) to (i1);
\draw [->](i12) to (i1);
\draw [->](i13) to (i12);
\draw [->](r1) to (i13);
\draw [->](r12) to (r1);

\draw [dashed, bend left=0] (j) to (join);
\draw [dashed, bend right=15] (rtx) to (rt);
\draw [dashed, bend right=20] (i22) to (leaf2);
\draw [dashed, bend right=40] (i13) to (leaf1);
\draw [dashed, bend left=22] (leaf0) to (l1);
\draw [dashed, bend left=22] (leaf0x) to (r1);
\end{tikzpicture}
\caption{Selected tables obtained by algorithm~$\dpa_{\algo{PRIM}}$ on tree decomposition~${\cal T}'$.}
\label{fig:running1}
\end{figure}

The following definition simplifies the presentation. At a node~$t$
and for a row~$\vec\tabval$ of the table $\ATab{\AlgS}[t]$, it yields
the rows in the tables of the children of~$t$ that were involved in
computing row~$\vec\tabval$ by algorithm~\AlgS.

\newcommand{\llangle}{\ensuremath{\langle\hspace{-2pt}\{\hspace{-0.2pt}}}
\newcommand{\rrangle}{\ensuremath{\}\hspace{-2pt}\rangle}}
\newcommand{\STab}{\ensuremath{\ATab{\AlgS}}}%

\begin{definition}[c.f.,~\cite{FichteEtAl17b}]\label{def:origin}
  Let $F$ be a formula, $\TTT=(T, \chi)$ be a tree decomposition of~$F$,
  $t$ be a node of~$T$ that has~$\ell$ children, and
  %
  %
  %
  $\tau_1, \ldots, \tau_{\ell}$ be the $\AlgS$-tables computed by
  $\dpa_\AlgS((F,\cdot),\TTT,\cdot)$ where
  $\children(t,T)=\langle t_1, \ldots, t_{\ell}\rangle$.
  Given a sequence~$\vec s=\langle s_1, \ldots, s_{\ell} \rangle$, we
  let
  $\llangle \vec s\rrangle \eqdef \langle \{s_1\}, \ldots,
  \{s_{\ell}\} \rangle$, for technical reasons.

  For a given $\AlgS$-row~$\vec u$, we define the originating
  $\AlgS$-rows of~$\vec u$ in node~$t$ by
  %
    $\orig(t,\vec \tabval) \eqdef \SB \vec s \SM \vec s \in 
    \tau_1 \times \cdots \times \tau_{\ell}, \tab{} =
    {\AlgS}(t,\chi(t),\prog_t, \cdot,\llangle \vec s\rrangle, 
    \cdot), \vec u \in \tab{} \SE.$ %
  %
  %
  We extend this to a $\AlgS$-table~$\sigma$ by
  $\origs(t,\sigma) \eqdef$ $\bigcup_{\vec u \in \sigma}\orig(t,\vec
    u).$
\end{definition}

\begin{remark}
  An actual implementation would not compute origins,
  but store and reuse them without side-effects to worst-case complexity
  during tree traversal.
\end{remark}


\begin{example}\label{ex:origins}
  Consider formula~$F$, tree decomposition~$\TTT'=(T,\chi)$, and
  tables $\tab{1}, \ldots, \tab{12}$ from Example~\ref{ex:sat}.  We
  focus
  on~$\vec{\tabval_{1.1}} = \langle J_{1.1} \rangle
  =\langle\emptyset\rangle$ of table~$\tab{1}$ of the leaf~$t_1$. The
  row~$\vec{\tabval_{1.1}}$ has no preceding row,
  since~$\type(t_1)=\leaf$. Hence, we have
  $\origse{\PRIM}(t_1,\vec{\tabval_{1.1}})=\{\langle \rangle\}$.
  The origins of row~$\vec{\tabval_{5.1}}$ of table~$\tab{5}$ are
  given by $\origse{\PRIM}(t_5,\vec{\tabval_{5.1}})$, which correspond
  to the preceding rows in table~$t_4$ that lead to
  row~$\vec{\tabval_{5.1}}$ of table~$\tab{5}$ when running
  algorithm~$\PRIM$,~i.e.,
  $\origse{\PRIM}(t_5,\vec{\tabval_{5.1}}) = \{\langle
  \vec{\tabval_{4.1}} \rangle, \langle\vec{\tabval_{4.4}}\rangle\}$.
  Observe that $\origse{\PRIM}(t_i,\vec\tabval)=\emptyset$ for any
  row~$\vec\tabval\not\in\tab{i}$.
  For node~$t_{11}$ of type~$\join$ and row~$\vec{\tabval_{11.1}}$, we
  obtain
  $\origse{\PRIM}(t_{11},\vec{\tabval_{11.1}}) =
  \{\langle\vec{\tabval_{6.2}},$ $\vec{\tabval_{10.1}} \rangle\}$
	(see Example~\ref{ex:sat}).
  %
  %
  More general, when using algorithm~\PRIM, at a node~$t$ of
  type~$\join$ with table~$\tau$ we have
  $\origse{\PRIM}(t, \vec u)=\{\langle \vec\tabval,
  \vec\tabval\rangle\}$ for 
	row~$\vec u \in \tau$. 
%
\end{example}

Definition~\ref{def:origin} talked about a top-down direction for rows
and their origins. In addition, we need definitions to talk about a
recursive version of these origins from a node~$t$ down to the leafs,
mainly to state properties for our algorithms.


\begin{definition}\label{def:extensions}
  Let $F$ be a formula, $\TTT=(T, \chi)$ be a tree decomposition with~$T=(N,\cdot,n)$, $t\in N$, $\ATab{\AlgS}[t']$ be obtained by $\dpa_\AlgS((F,\cdot),\TTT,\cdot)$
  for each node~$t'$ of the \emph{induced sub-tree~$T[t]$ rooted at~$t$}, and $\vec u$ be a row of $\ATab{\AlgS}[t]$.

  An \emph{extension below~$t$} is a set of pairs where a pair consists
  of a node~$t'$ of~$T[t]$ and a row~$\vec v$ of $\ATab{\AlgS}[t']$
  and the cardinality of the set equals the number of nodes in the
  sub-tree~$T[t]$. We define the family of \emph{extensions below~$t$}
  recursively as follows.  If $t$ is of type~\leaf, then
  $\Ext_{\leq t}(\vec u) \eqdef \{\{\langle t,\vec u\rangle\}\}$;
  otherwise
  $\Ext_{\leq t}(\vec u) \eqdef \bigcup_{\vec v \in \origs(t,\vec u)}
  \big\SB\{\langle t,\vec u\rangle\}\cup X_1 \cup \ldots \cup X_\ell
  \SM X_i\in\Ext_{\leq t_i}({\vec v}_{(i)})\big\SE$ 
  for the~$\ell$ children~$t_1, \ldots, t_\ell$ of~$t$.
  %
  We extend this notation for a $\AlgS$-table~$\sigma$ by
  $\Ext_{\leq t}(\sigma)\eqdef \bigcup_{\vec u\in\sigma} \Ext_{\leq
    t}(\vec u)$.  Further, we
  let~$\Exts \eqdef \Ext_{\leq n}(\ATab{\AlgS}[n])$. 

\end{definition}

If we would construct all extensions below the root~$n$, it allows us
to also obtain all models of a formula~$F$.  To this end, we state the following definition.

\begin{definition}\label{def:satext}
  Let $F$ be a formula, $\TTT=(T, \chi)$ be a tree decomposition of~$F$, $t$
  be a node of $T$, and $\sigma \subseteq \ATab{\AlgS}[t]$ be a set
  of~$\AlgS$-rows that have been computed by
  $\dpa_\AlgS((F,\cdot),\TTT,\cdot)$ at~$t$.
  We define 
  the \emph{satisfiable
    extensions below~$t$} for~$\sigma$ by
  $\PExt_{\leq t}(\sigma)\eqdef \bigcup_{\vec u\in\sigma} \SB X \SM X
    \in \Ext_{\leq t}(\vec u), X \subseteq Y, Y \in \Exts\SE.$
\end{definition}

\begin{observation}
Let~$F$ be a formula,~${\cal T}$ be a tree decomposition with root~$n$ of~$F$. Then, $\PExt_{\leq n}(\ATab{\AlgS}[t]) = \Exts$.
\end{observation}

Next, we define an auxiliary notation that gives us a way to
reconstruct interpretations from families of extensions.


\begin{definition}\label{def:iextensions}
 Let~$(F, P)$ be an instance of \PMC, $\TTT=(T, \chi)$ be a tree decomposition
  of~$F$, $t$ be a node of~$T$. Further, let $E$ be a family of extensions below~$t$, and $P$ be a set of projection variables. We
  define the \emph{set~$I(E)$ of interpretations} of~$E$ by
  $I(E) \eqdef \big\SB \bigcup_{\langle \cdot, \vec u \rangle \in X} I(\vec u)
  \mid X \in E \big\SE$
  and the set~$I_P(E)$ of \emph{projected interpretations} by
  $I_P(E) \eqdef \big\SB \bigcup_{\langle \cdot, \vec u \rangle \in X} I(\vec
  u) \cap P \mid X \in E \big\SE$.

\end{definition}

\begin{example}
  Consider again formula~$F$ and tree decomposition~${\cal T}'$ with
  root~$n$ of~$F$ from Example~\ref{ex:sat}.
  Let~$X=\{\langle t_{12}, \langle\emptyset\rangle\rangle, \langle t_{11},
  \langle\{a\}\rangle\rangle,$
  $\langle t_6, \langle\{a\}\rangle \rangle, \langle t_5, \langle\{a,b\}\rangle\rangle,$ $\langle
  t_4,\hspace{-0.1em} \langle\{a,b\}\rangle\rangle,$
  $\langle t_3,\hspace{-0.1em} \langle\{a\}\rangle \rangle, \langle t_2,\hspace{-0.1em} \langle\{a\}\rangle\rangle, \langle t_1,\hspace{-0.1em}
  \langle\emptyset \rangle\rangle, \langle t_{10},\hspace{-0.1em} \langle\{a\}\rangle\rangle,\langle t_9,\hspace{-0.1em}
  $ $\langle\{a,p_2\}\rangle\rangle, \langle t_8,\hspace{-0.1em} \langle\{p_2\}\rangle\rangle,$ $\langle t_7, \langle\emptyset
  \rangle\rangle \}$ be an extension below~$n$.  Observe that~$X\in\Exts$ and
  that Figure~\ref{fig:running1} highlights those rows of 
  tables for nodes~$t_{12},t_{11},t_9,t_5,t_4$ and~$t_1$ that also
  occur in~$X$ (in yellow). Further, $I(\{X\})=\{a,b,p_2\}$ computes
  the corresponding model of~$X$, and $I_P(\{X\}) = \{p_2\}$ derives
  the projected model of~$X$.  $I(\Exts)$ refers to the set of
  models of~$F$, whereas~$I_P(\Exts)$ is the set of projected models of~$F$.
\end{example}

\section{Counting Projected Models by Dynamic Programming}\label{sec:projmodelcounting}

\longversion{
\begin{figure}[t]
\centering
\includegraphics[scale=0.8]{figure_projection.pdf}
\caption{Algorithm~$\mdpa{\AlgS}$ consists of~$\dpa_\AlgS$
  and~$\dpa_\PROJ$. 
}
\label{fig:multiarch}
\end{figure}%
}

In this section, we introduce the dynamic programming
algorithm~\mdpa{\AlgS} to solve the projected model counting problem
(\PMC) for Boolean formulas.
Our algorithm traverses the tree decomposition twice following a
multi-pass dynamic programming paradigm~\cite{FichteEtAl17b}.
\longversion{Figure~\ref{fig:multiarch} illustrates the steps of our
algorithm~\mdpa{\AlgS}.}
Similar to the previous section\longversion{ (Figure~\ref{fig:framework})}, we
construct a graph representation and heuristically compute a tree
decomposition of this graph. Then, we run $\dpa_\AlgS$ (see
Listing~\ref{fig:dpontd}) in Step~3a as first traversal. Step~3a can
also be seen as a preprocessing step for projected model counting,
from which we immediately know whether the problem has a
solution. Afterwards we remove all rows from the \AlgS-tables which
cannot be extended to a solution for the \SAT problem (\emph{``Purge
  non-solutions''}).
In other words, we keep only rows~$\vec u$ in table~$\ATab{\AlgS}[t]$
at node~$t$ if its interpretation~$I(\vec u)$ can be extended to a
model of~$F$, more formally, $(t,\vec u)\in X$ for some~$X \in \PExt_{\leq t}(\ATab{\AlgS}[t])$.
%
%
%
Thereby, we avoid redundancies and can simplify the description of our
next step, since we then only have to consider (parts of) models.
In Step~3b ($\dpa_\PROJ$), we traverse the tree decomposition a second
time to count projections of interpretations of rows in
$\AlgS$-tables.
%
%
In the following, we only describe the table algorithm~$\PROJ$, since
the traversal in $\dpa_\PROJ$ is the same as before.
For \PROJ, 
a row at a node~$t$ is a pair $\langle\sigma, c \rangle$ where $\sigma$ is a
$\AlgS$-table, in particular, a subset of $\ATab{\AlgS}[t]$
computed by $\dpa_\AlgS$, and $c$ is a non-negative integer.
In fact, we store in integer~$c$ a count that expresses the number of
``all-overlapping'' solutions ($\ipmc$), whereas in the end we aim for the
projected model count ($\pmc$), clarified in the following. 


%

\begin{definition}\label{def:pmc}
  Let $F$ be a formula, $\TTT=(T, \chi)$ be a tree decomposition of~$F$,
  $t$ be a node of~$T$, $\sigma \subseteq \ATab{\AlgS}[t]$ be a set
  of~$\AlgS$-rows that have been computed by
  $\dpa_\AlgS((F,\cdot),\TTT,\cdot)$ at node~$t$ in~$T$.
  Then, the \emph{projected model count} $\pmc_{\leq t}(\sigma)$ of
  $\sigma$ below~$t$ is the size of the union over projected
  interpretations of the satisfiable extensions of~$\sigma$ below~$t$,
  formally,
  $\pmc_{\leq t}(\sigma) \eqdef \Card{\bigcup_{\vec u\in\sigma}
    I_P(\PExt_{\leq t}(\{\vec u\}))}$.

  The \emph{intersection projected model count}
  $\ipmc_{\leq t}(\sigma)$ of $\sigma$ below~$t$ is the size of the
  intersection over projected interpretations of the satisfiable
  extensions of~$\sigma$ below~$t$,~i.e.,
  $\ipmc_{\leq t}(\sigma) \eqdef \Card{\bigcap_{\vec u\in\sigma}
    I_P(\PExt_{\leq t}(\{\vec u\}))}$.
\end{definition}

The next definitions provide central notions for grouping rows of
tables according to the given projection of variables.

\newcommand{\RRR}{\ensuremath{\mathcal{R}}}


\begin{definition}
  Let $(F,P)$ be an instance of \PMC and $\sigma$ be a $\AlgS$-table.
  We define the relation~$\bucket \subseteq \sigma \times \sigma$ to
  consider equivalent rows with respect to the projection of its
  interpretations by 
  $\bucket \eqdef \SB (\vec u,\vec v) \SM \vec u, \vec v \in \sigma,
  \restrict{I(\vec u)}{P} = \restrict{I(\vec v)}{P}\SE.$
\end{definition}

\begin{observation}\label{obs:relation}
  The relation~$\bucket$ is an equivalence relation.
\end{observation}

\begin{definition}
  Let~$\tau$ be a $\AlgS$-table and $\vec u$ be a row of $\tau$.  The
  relation~$\bucket$ induces equivalence classes~$[\vec u]_P$ on the
  $\AlgS$-table~$\tau$ in the usual way,~i.e.,
  $[\vec u]_P = \SB \vec v \SM \vec v \bucket \vec u,\vec v \in
  \tau\}$~\cite{Wilder12a}.
  We denote by~$\buckets_P(\tau)$ the set of equivalence classes
  of~$\tau$,~i.e.,
  $\buckets_P(\tau) \eqdef\, (\tau / \bucket) = \SB [\vec u]_P \SM
  \vec u \in \tau\SE$.
  Further, we define the set $\subbuckets_P(\tau)$ of all
  sub-equivalence classes of~$\tau$ by
  $\subbuckets_P(\tau) \eqdef \SB S \SM \emptyset \subsetneq S
  \subseteq B, B \in \buckets_P(\tau)\SE$.

\end{definition}

\begin{example}\label{ex:equiv}
  Consider again formula~$F$ and set~$P$ of projection variables from
  Example~\ref{ex:running0} and tree
  decomposition~$\mathcal{T}'=(T,\chi)$ and $\AlgS$-table~$\tab{4}$
  from Figure~\ref{fig:running1}.
  %
  %
  We have $\vec{ u_{4.1}} =_P \vec{ u_{4.2}}$ and
  $\vec{ u_{4.4}} =_P \vec{ u_{4.5}}$.  We obtain the
  set~$\tab{4}/\bucket$ of equivalence classes of $\tab{4}$
  by~$\buckets_P(\tab{4})=\{\{\vec{ u_{4.1}}, \vec{ u_{4.2}}, \vec{
    u_{4.3}}\}, \{\vec{ u_{4.4}},$
  $ \vec{ u_{4.5}}, \vec{ u_{4.6}}\}\}$.
\end{example}

Since $\PROJ$ stores a counter in $\PROJ$-tables together with a
$\AlgS$-table, we need an auxiliary definition that given
$\AlgS$-table allows us to select the respective counts from a
$\PROJ$-table.
Later, we use the definition in the context of looking up the already
computed projected counts for tables of \emph{children} of a given
node.

\begin{definition}\label{def:childpcnt}
  Given a $\PROJ$-table~$\iota$ and a $\AlgS$-table~$\sigma$ we define
  the \emph{stored $\ipmc$} for all rows of~$\sigma$ in~$\iota$ by
  $\sipmc(\iota, \sigma) \eqdef \sum_{\langle \sigma, c\rangle \in
    \iota} c.$
  %
  Later, we apply this to rows from several origins.
  Therefore, for a
  sequence~$s$ of $\PROJ$-tables of length $\ell$ and a set~$O$ of
  sequences of $\AlgS$-rows where each sequence is of length~$\ell$,
  we let
  $\sipmc(s, O)=\prod_{i \in \{1, \ldots,
    \ell\}}\sipmc(s_{(i)},O_{(i)}).$
\end{definition}

When computing $\sipmc$ in Definition~\ref{def:childpcnt}, we select
the $i$-th position of the sequence together with sets of the $i$-th
position from the set of sequences. We need this somewhat technical
construction, since later at node~$t$ we apply this definition to
$\PROJ$-tables of children of~$t$ and origins of subsets of
$\AlgS$-tables. There, we may simply have several children if the node
is of type~$\join$ and hence we need to select from the right
children.

Now, we are in position to give a core definition for our algorithm
that solves \PMC.
Intuitively, when we are at a node~$t$ in the Algorithm~$\dpa_\PROJ$
we already computed all tables $\ATab{\AlgS}$ by $\dpa_\AlgS$
according to Step~3a, purged non-solutions, and computed
$\ATab{\PROJ}[t']$ for all nodes~$t'$ below~$t$ and in particular the
$\PROJ$-tables~$\Tab{}$ of the children of~$t$.
Then, we compute the projected model count of a subset~$\sigma$ of the
$\AlgS$-rows in~$\ATab{\AlgS}[t]$, which we formalize in the following
definition, by applying the generalized inclusion-exclusion principle
to the stored projected model count of origins.
%

\begin{definition}\label{def:pcnt}
  Let $(F,P)$ be an instance of \PMC, $\TTT=(T, \chi)$ be a tree
  decomposition of~$F$,
  $\ATab{\AlgS}[s]$ be the $\AlgS$-tables computed by
  $\dpa_\AlgS((F,\cdot),\TTT,\cdot)$ for every node~$s$ of $T$.
  Further, let $t$ be a node of~$T$ with $\ell$ children,
  $\Tab{} = \langle \ATab{\PROJ}[t_1], \ldots,
  \ATab{\PROJ}[t_{\ell}]\rangle$ be the sequence of $\PROJ$-tables
  computed by $\dpa_\PROJ((F,P),\TTT,\ATab{\AlgS})$ where
  $\children(t,T)=\langle t_1, \ldots, t_{\ell}\rangle$, and
  $\sigma \subseteq \ATab{\AlgS}[t]$ be a table.
  We define the \emph{(inductive) projected model count} of $\sigma$: \vspace{-0.75em}
  \begin{align*}
    \pcnt(t,\sigma, \Tab{}) \eqdef & %
                                   \sum_{\emptyset \subsetneq O \subseteq {\origs(t,\sigma)}} (-1)^{(\Card{O} - 1)} \cdot
                                 & \sipmc(\Tab{}, O).%
  \end{align*}

\end{definition}



Vaguely speaking, $\pcnt$ determines the $\AlgS$-origins of the
set~$\sigma$ of rows, goes over all subsets of these origins and looks up
the stored counts ($\sipmc$) in the $\PROJ$-tables of the children
of~$t$. Example~\ref{ex:pcnt} provides an idea on how to compute the
projected model count of tables of our running example using~$\pcnt$.

\begin{example}\label{ex:pcnt}
  The function defined in Definition~\ref{def:pcnt} allows us to
  compute the projected count for a given~\AlgS-table. Therefore,
  consider again formula~$F$ and tree decomposition~$\TTT'$ from
  Example~\ref{ex:running1} and Figure~\ref{fig:running1}. Say we want
  to compute the projected count $\pcnt(t_5,\{\vec{ u_{5.4}}\}, \Tab{})$
  where
  $\Tab{}\eqdef\allowdisplaybreaks[4] \big\SB \langle \{\vec{
    u_{4.3}}\}, 1\rangle,$
  $\langle \{\vec{ u_{4.6}}\},1\rangle\big\SE$ for
  row~$\vec{ u_{5.4}}$ of table $\tab{5}$. Note that~$t_5$ has
  $\ell=1$ child nodes~$\langle t_4 \rangle$ and therefore the product
  of Definition~\ref{def:childpcnt} consists of only one
  factor. Observe that
  $\origse{\PRIM}(t_5, \vec{ u_{5.4}}) = \{\langle\vec{ u_{4.3}}\rangle, \langle\vec{
    u_{4.6}}\rangle\}$. Since the rows~$\vec{ u_{4.3}}$ and~$\vec{ u_{4.6}}$
  do not occur in the same \PRIM-table of~\Tab{}, only the value of
  $\sipmc$ for the two singleton origin sets~$\{\langle\vec{ u_{4.3}}\rangle\}$ and
  $\{\langle\vec{ u_{4.6}}\rangle\}$ is non-zero; for the remaining set of origins we have zero. Hence, we obtain $\pcnt(t_5,\{\vec{ u_{5.4}}\}, \Tab{})=2$.
\end{example}

\noindent Before we present algorithm~$\PROJ$
(Listing~\ref{fig:dpontd3}), we give a definition that allows us at a
certain node~$t$ to compute the intersection
$\pmc$ for a given~$\AlgS$-table~$\sigma$ by computing the $\pmc$ (using stored $\ipmc$ values from
$\PROJ$-tables for children of~$t$), and subtracting and adding~$\ipmc$ values for subsets~$\emptyset\subsetneq\rho\subsetneq\sigma$ accordingly.

\begin{definition}\label{def:ipmc}
  Let $\TTT=(T,\cdot)$ be a tree decomposition, $t$ be a node of~$T$,
  $\sigma$ be $\AlgS$-table, and $\Tab{}$ be a sequence of tables. 
  Then, we define the \emph{(recursive) $\ipmc$} of~$\sigma$ as follows:
  \springerversion{\vspace{-1.5em}}
  \begin{align*}
    \icnt(t,\sigma,\Tab{})\eqdef
    \begin{cases} %
      1, \text{ if } \type(t) = \leaf,\\
      \big|\pcnt(t,\sigma, \Tab{})\;+ \\ \quad\sum_{\emptyset\subsetneq\rho\subsetneq\sigma}(-1)^{\Card{\rho}}
        \cdot \ipmc(t,\rho, \Tab{})\big|, \text{otherwise.}
    \end{cases}
  \end{align*}
\end{definition}

\noindent In other words, if a node is of type~$\leaf$ the $\ipmc$ is one, since
by definition of a tree decomposition the bags of nodes of
type~$\leaf$ contain only one projected interpretation (the empty set).
%
%
%
%
Otherwise, using Definition~\ref{def:pcnt}, we are able to compute the
$\ipmc$ for a given $\AlgS$-table~$\sigma$, which is by construction the same
as $\ipmc_{\leq t}(\sigma)$ (c.f.\ proof of
Theorem~\ref{thm:correctness} later).
In more detail, we want to compute for a $\AlgS$-table~$\sigma$ its
$\ipmc$ that represents ``all-overlapping'' counts of~$\sigma$ with
respect to set~$P$ of projection variables, that is,
$\ipmc_{\leq t}(\sigma)$. Therefore, for $\ipmc$, we rearrange the
inclusion-exclusion principle.
To this end, we take $\pcnt$, which computes the ``non-overlapping''
count of~$\sigma$ with respect to~$P$, by once more exploiting the
inclusion-exclusion principle on $\AlgS$-origins of~$\sigma$ (as already discussed) such that
we count every projected model only once. Then we have to alternately subtract and add $\ipmc$ values for strict subsets~$\rho$ of~$\sigma$, accordingly.

Finally, Listing~\ref{fig:dpontd3} presents table algorithm~\PROJ,
which stores for given node~$t$ a \PROJ-table consisting of every
sub-bucket of the given table~$\ATabs{\AlgS}{t}$ together with its
$\ipmc$ (as presented above).




  
\begin{algorithm}[t]
  \KwData{ %
    Node~$t$, set~$P$ of projection variables, $\Tab{}$, and 
    $\ATab{\AlgS}$. 
    %
  }%
  \KwResult{Table~$\iota_{t}$ consisting of pairs~$\langle \sigma,
    c\rangle$, where $\sigma \subseteq \ATabs{\AlgS}{t}$ and $c \in
    \NAT$.\hspace{-5em}
  } %
  $\makebox[0em]{}\iota_{t} \leftarrow \big\SB\langle \sigma,
  \icnt(t,\sigma,\Tab{}) \rangle \big{|}\, \sigma \in
  \subbuckets_P(\ATabs{\AlgS}{t})\big\SE\hspace{-5em}$ \;
  \Return{$\iota_{t}$}
  \vspace{-0.15em}
  \caption{Table algorithm $\PROJ(t, \cdot, \cdot, P, \Tab{},
    \ATab{\AlgS})$.}
  \label{fig:dpontd3}
\end{algorithm}


%
%


\begin{example}
  Recall instance~$(F,P)$, tree decomposition~$\TTT'$, and 
  tables~$\tab{1}$, $\ldots$, $\tab{12}$ from
  Example~\ref{ex:running0}, \ref{fig:running1}, and
  Figure~\ref{fig:running1}. Figure~\ref{fig:running2} depicts
  selected tables of~$\iota_1, \ldots \iota_{12}$ obtained after
  running~$\dpa_\PROJ$ for counting projected
  interpretations. 
  We assume numbered rows,~i.e., 
  row $i$ in table $\iota_t$ corresponds to
  $\vec{v_{t.i}} = \langle \sigma_{t.i}, c_{t.i} \rangle$.
  Note that for some nodes~$t$, there are rows among
  different~$\PRIM$-tables that occur in~$\Ext_{\leq t}$, but not
  in~$\PExt_{\leq t}$. These rows are removed
  during purging. In fact,
  rows~$\vec{\tabval_{4.1}}, \vec{\tabval_{4.2}}$,
  and~$\vec{\tabval_{4.4}}$ do not occur in table~$\iota_4$. Observe
  that purging is a crucial trick here that avoids to correct
  stored counters~$c$ by backtracking whenever a certain row of a
  table has no succeeding row in the parent table.
  
  Next, we discuss selected rows obtained
  by~$\dpa_\PROJ((F,P),\TTT',\ATab{\PRIM})$. Tables $\iota_1$,
  $\ldots$, $\iota_{12}$ that are computed at the respective nodes of
  the tree decomposition are shown in Figure~\ref{fig:running2}.
  Since~$\type(t_1)= \leaf$, we have
  $\iota_1=\langle\{\langle \emptyset \rangle \}, 1\rangle$.
  Intuitively, up to node~$t_1$ the
  \AlgS-row~$\langle\emptyset\rangle$ belongs to~$1$ bucket.
  Node~$t_2$ introduces variable~$a$, which results in
  table~$\iota_2\eqdef\big\SB\langle \{\langle \{a\} \rangle \},
  1\rangle\big\SE$. Note that the $\PRIM$-row
  $\langle\emptyset\rangle$ is subject to purging.
  Node~$t_3$ introduces~$p_1$ and node~$t_4$ introduces~$b$.
  Node~$t_5$ removes projected variable~$p_1$.  The
  row~$\vec{v_{5.2}}$ of $\PROJ$-table~$\iota_5$ has already been
  discussed in Example~\ref{ex:pcnt} and row~$\vec{v_{5.1}}$ works
  similar.
  For row~$\vec{v_{5.3}}$ we compute the
  count~$\ipmc(t_5,\{\vec{\tabval_{5.2}},\vec{\tabval_{5.4}}\},
  \langle \iota_4\rangle)$ by means of~$\pcnt$. Therefore, take
  for~$\rho$ the sets~$\{\vec{\tabval_{5.2}}\}$,
  $\{\vec{\tabval_{5.4}}\}$, and
  $\{\vec{\tabval_{5.2}},\vec{\tabval_{5.4}}\}$.
  For the singleton sets, we simply have
  $\pmc(t_5,\{\vec{\tabval_{5.2}}\}, \langle \iota_4\rangle) = \ipmc(t_5,\{\vec{\tabval_{5.2}}\}, \langle \iota_4\rangle) = c_{5.1}
  = 1$ and
  $\pmc(t_5,\{\vec{\tabval_{5.4}}\}, \langle \iota_4\rangle )= \ipmc(t_5,\{\vec{\tabval_{5.4}}\}, \langle \iota_4\rangle ) =
  c_{5.2}=2$.
  To compute
  $\pmc(t_5,\{\vec{\tabval_{5.2}},\vec{\tabval_{5.4}}\}, \langle
  \iota_4\rangle)$ following Definition~\ref{def:pcnt}, take for~$O$
  the sets~$\{\vec{u_{4.5}}\}$, $\{\vec{u_{4.3}}\}$, and
  $\{\vec{u_{4.6}}\}$ into account, since all other non-empty subsets
  of origins of~$\vec{\tabval_{5.2}}$ and~$\vec{\tabval_{5.4}}$
  in~$\iota_4$ do not occur in~$\iota_4$.
  Then, we take the sum over the values
  $\sipmc(\langle t_4\rangle, \{\langle\vec{\tabval_{4.5}}\rangle\})=1$,
  $\sipmc(\langle t_4\rangle, \{\langle\vec{\tabval_{4.3}}\rangle\})=1$, and
  $\sipmc(\langle t_4\rangle, \{\langle\vec{\tabval_{4.6}}\rangle \})$ $=1$; and
  subtract~$\sipmc(\langle t_4\rangle, $
  $\{\langle\vec{\tabval_{4.5}}\rangle, \langle\vec{\tabval_{4.6}}\rangle\})=1$. 
  Hence, 
  $\pmc(t_5,\{\vec{\tabval_{5.2}},\vec{\tabval_{5.4}}\},$ $\langle
  \iota_4\rangle)=2$.
  In order to
  compute~$\ipmc(t_5,\{\vec{\tabval_{5.2}},\vec{\tabval_{5.4}}\},
  \langle \iota_4\rangle) = | %
  \pmc(t_5,\{\vec{\tabval_{5.2}},\vec{\tabval_{5.4}}\},$ $\langle
  \iota_4\rangle)
  - \ipmc(t_5,\{\vec{\tabval_{5.2}}\}, \langle \iota_4\rangle) 
  - \ipmc(t_5,\{\vec{\tabval_{5.4}}\}, \langle \iota_4\rangle) 
  | = |2 -1 - 2| =|-1| = 1$.
  Hence, $c_{5.3} = 1$ represents the number of projected models, both
  rows~$\vec{u_{5.2}}$ and~$\vec{u_{5.4}}$ have in common. We then use it for table~$t_6$.

  For node~$t_{11}$ of type~$\join$ one simply in addition multiplies
  stored $\sipmc$ values for \AlgS-rows in the two children
  of~$t_{11}$ accordingly (see Definition~\ref{def:childpcnt}).
  In the end, the projected model count of~$F$ corresponds to~$\sipmc(\iota_{12},\cdot)=4$. 
\end{example}

\begin{figure}[t]
\centering
\begin{tikzpicture}[node distance=0.5mm]
\tikzset{every path/.style=thick}

\node (l1) [stdnode,label={[tdlabel, xshift=0em,yshift=+0em]right:${t_1}$}]{$\emptyset$};
\node (i1) [stdnode, above=of l1, label={[tdlabel, xshift=0em,yshift=+0em]right:${t_2}$}]{$\{a\}$};
\node (i12) [stdnode, above=of i1, label={[tdlabel, xshift=0em,yshift=+0em]right:${t_3}$}]{$\{a,p_1\}$};
\node (i13) [stdnode, above=of i12, label={[tdlabel, xshift=0em,yshift=+0em]right:${t_4}$}]{$\{a,b,p_1\}$};
\node (r1) [stdnode, above=of i13, label={[tdlabel, xshift=0em,yshift=+0em]right:${t_5}$}]{$\{a,b\}$};
\node (r12) [stdnode, above=of r1, label={[tdlabel, xshift=0em,yshift=+0em]right:${t_6}$}]{$\{a\}$};
\node (l2) [stdnode, right=2.5em of i12, label={[tdlabel, xshift=0em,yshift=+0em]left:${t_7}$}]{$\emptyset$};
\node (i2) [stdnode, above=of l2, label={[tdlabel, xshift=0em,yshift=+0em]left:${t_8}$}]{$\{p_2\}$};
\node (i22) [stdnode, above=of i2, label={[tdlabel, xshift=0em,yshift=+0em]left:${t_9}$}]{$\{a,p_2\}$};
\node (r2) [stdnode, above=of i22, label={[tdlabel, xshift=0em,yshift=+0em]left:${t_{10}}$}]{$\{a\}$};
\node (j) [stdnode, above left=of r2, yshift=-0.25em, label={[tdlabel, xshift=0em,yshift=+0.15em]right:${t_{11}}$}]{$\{a\}$};
\node (rt) [stdnode,ultra thick, above=of j, label={[tdlabel, xshift=0em,yshift=+0em]right:${t_{12}}$}]{$\emptyset$};
\node (label) [font=\scriptsize,left=of rt]{${\cal T}'$:};
\node (leaf1) [stdnode, left=1.6em of i1, yshift=0.5em, label={[tdlabel, xshift=3.25em,yshift=1em]below right:$\iota_{4}$}]{%
	\begin{tabular}{l@{\hspace{0.0em}}r}%
		\multicolumn{1}{l}{$\langle \tuplecolor{\outputPredColor}{\sigma_{4.i}}, $}&\multicolumn{1}{r}{$\tuplecolor{\statePredColor}{c_{4.i}} \rangle$}\\
		\hline\hline
		$\langle\tuplecolor{\outputPredColor}{\{\langle}\tuplecolor{\inputPredColor}{\{a,b\}}\tuplecolor{\outputPredColor}{\rangle\}}, $&$\tuplecolor{\statePredColor}{1}\rangle$\\\specialrule{.1em}{.05em}{.05em}

		$\langle\tuplecolor{\outputPredColor}{\{\langle}\tuplecolor{\inputPredColor}{\{a,p_1\}}\tuplecolor{\outputPredColor}{\rangle\}}, $&$\tuplecolor{\statePredColor}{1}\rangle$\\\hline
		$\langle\tuplecolor{\outputPredColor}{\{\langle}\tuplecolor{\inputPredColor}{\{a,b,p_1\}}\tuplecolor{\outputPredColor}{\rangle\}}, $&$\tuplecolor{\statePredColor}{1}\rangle$\\\hline
		$\langle\tuplecolor{\outputPredColor}{\{\langle}\tuplecolor{\inputPredColor}{\{a,p_1\}}\tuplecolor{\outputPredColor}{\rangle, } \tuplecolor{\outputPredColor}{\langle} \tuplecolor{\inputPredColor}{\{a,b,p_1\}}\tuplecolor{\outputPredColor}{\rangle\}}, $&$\tuplecolor{\statePredColor}{1}\rangle$
	\end{tabular}%
};
\node (leaf1b) [stdnodenum,left=of leaf1,xshift=0.6em,yshift=0pt]{%
	\begin{tabular}{c}%
		\multirow{1}{*}{$i$}\\ 
		\hline\hline
		$1$ \\\specialrule{.1em}{.05em}{.05em}
		$2$ \\\hline
		$3$ \\\hline
		$4$
	\end{tabular}%
};
\node (leaf0x) [stdnode, left=-5.25em of leaf1b, yshift=6.5em, label={[tdlabel, xshift=4em,yshift=0.25em]above left:$\iota_{5}$}]{%
	\begin{tabular}{l@{\hspace{0.0em}}r}%
		\multicolumn{1}{l}{$\langle \tuplecolor{\outputPredColor}{\sigma_{5.i}}, $}&\multicolumn{1}{r}{$\tuplecolor{\statePredColor}{c_{5.i}} \rangle$}\\
		\hline\hline
		$\langle\tuplecolor{\outputPredColor}{\{\langle}\tuplecolor{\inputPredColor}{\{a\}}\tuplecolor{\outputPredColor}{\rangle\}}, $&$\tuplecolor{\statePredColor}{1}\rangle$\\\hline
		$\langle\tuplecolor{\outputPredColor}{\{\langle}\tuplecolor{\inputPredColor}{\{a,b\}}\tuplecolor{\outputPredColor}{\rangle\}}, $&$\tuplecolor{\statePredColor}{2}\rangle$\\\hline	
		$\langle\tuplecolor{\outputPredColor}{\{\langle}\tuplecolor{\inputPredColor}{\{a\}}\tuplecolor{\outputPredColor}{\rangle, } \tuplecolor{\outputPredColor}{\langle} \tuplecolor{\inputPredColor}{\{a,b\}}\tuplecolor{\outputPredColor}{\rangle\}}, $&$\tuplecolor{\statePredColor}{1}\rangle$\\
	\end{tabular}%
};
\node (leaf0b) [stdnodenum,left=of leaf0x,xshift=0.6em,yshift=0pt]{%
	\begin{tabular}{c}%
		\multirow{1}{*}{$i$}\\ 
		\hline\hline
		$1$ \\\hline
		$2$ \\\hline
		$3$ \\
	\end{tabular}%
};
\node (leaf2b) [stdnodenum,right=3em of j,xshift=-0.25em,yshift=-5.75em]  {%
	\begin{tabular}{c}%
		\multirow{1}{*}{$i$}\\ 
		\hline\hline
		$1$\\\specialrule{.1em}{.05em}{.05em}
		$2$\\
	\end{tabular}%
};
\node (leaf2) [stdnode,right=-0.4em of leaf2b, label={[tdlabel, xshift=0em,yshift=-0.25em]below:$\iota_{9}$}]  {%
	\begin{tabular}{l@{\hspace{0.0em}}r}%
		\multirow{1}{*}{$\langle \tuplecolor{\outputPredColor}{\sigma_{9.i}},$}& $\tuplecolor{\statePredColor}{c_{9.i}} \rangle$\\ 
		\hline\hline
		$\langle \tuplecolor{\outputPredColor}{\{\langle}\tuplecolor{\inputPredColor}{\{a\}}\tuplecolor{\outputPredColor}{\rangle\}},$ &$\tuplecolor{\statePredColor}{1}\rangle$\\\specialrule{.1em}{.05em}{.05em}
		$\langle \tuplecolor{\outputPredColor}{\{\langle}\tuplecolor{\inputPredColor}{\{a,p_2\}}\tuplecolor{\outputPredColor}{\rangle\}}, $& $\tuplecolor{\statePredColor}{1}\rangle$\\
	\end{tabular}%
};
\coordinate (middle) at ($ (leaf1.north east)!.5!(leaf2.north west) $);
\node (join) [stdnode,left=0.75em of r12, yshift=2.5em, label={[tdlabel, xshift=0.1em,yshift=+0.5em]right:$\iota_{{11}}$}] {%
	\begin{tabular}{l@{\hspace{0.0em}}r}%
		\multirow{1}{*}{$\langle \tuplecolor{\outputPredColor}{\sigma_{11.i}},$}& $\tuplecolor{\statePredColor}{c_{11.i}} \rangle$\\
		\hline\hline
		$\langle \tuplecolor{\outputPredColor}{\{\langle}\tuplecolor{\inputPredColor}{\{a\}}\tuplecolor{\outputPredColor}{\rangle\}},$ & $\tuplecolor{\statePredColor}{4}\rangle$\\
	\end{tabular}
};
\node (joinb) [stdnodenum,left=-0.45em of join] {%
	\begin{tabular}{c}
		\multirow{1}{*}{$i$}\\
		\hline\hline
		$1$ \\
	\end{tabular}%
};
\node (leaf0n) [stdnodenum,yshift=0.5em, right=2.5em of l1] {%
	\begin{tabular}{c}%
		\multirow{1}{*}{$i$}\\ 
		\hline\hline
		$1$
	\end{tabular}%
};
\node (leaf0) [stdnode,right=-0.5em of leaf0n, label={[tdlabel, xshift=0.1em,yshift=0.15em]right:$\iota_{1}$}] {%
	\begin{tabular}{l@{\hspace{0.0em}}r}%
		\multicolumn{1}{l}{$\langle \tuplecolor{\outputPredColor}{\sigma_{1.i}}, $}&$\tuplecolor{\statePredColor}{c_{1.i}} \rangle$\\
		\hline\hline
		$\langle\tuplecolor{\outputPredColor}{\{\langle} \tuplecolor{\inputPredColor}{\emptyset}\tuplecolor{\outputPredColor}{\rangle\}}, $&$\tuplecolor{\statePredColor}{1}\rangle$
	\end{tabular}%
};
\node (joinrrt) [stdnode,right=6.5em of r12, yshift=2.75em, label={[tdlabel, xshift=0.1em,yshift=+0.25em]right:$\iota_{{12}}$}] {%
	\begin{tabular}{l@{\hspace{0.0em}}r}%
		\multirow{1}{*}{$\langle \tuplecolor{\outputPredColor}{\sigma_{12.i}},$} & $\tuplecolor{\statePredColor}{c_{12.i}} \rangle$\\
		\hline\hline
		$\langle \tuplecolor{\outputPredColor}{\{\langle}\tuplecolor{\inputPredColor}{\emptyset}\tuplecolor{\outputPredColor}{\rangle\}},$ &$\tuplecolor{\statePredColor}{4}\rangle$\\
	\end{tabular}
};
\node (joinrbrt) [stdnodenum,left=-0.45em of joinrrt] {%
	\begin{tabular}{c}
		\multirow{1}{*}{$i$}\\
		\hline\hline
		$1$ \\
	\end{tabular}%
};
\node (joinr) [stdnode,right=6.5em of r12, yshift=-0.2em, label={[tdlabel, xshift=0.1em,yshift=+0.25em]right:$\iota_{{10}}$}] {%
	\begin{tabular}{l@{\hspace{0.0em}}r}%
		\multirow{1}{*}{$\langle \tuplecolor{\outputPredColor}{\sigma_{10.i}},$} & $\tuplecolor{\statePredColor}{c_{10.i}} \rangle$\\
		\hline\hline
		$\langle \tuplecolor{\outputPredColor}{\{\langle}\tuplecolor{\inputPredColor}{\{a\}}\tuplecolor{\outputPredColor}{\rangle\}},$ &$\tuplecolor{\statePredColor}{2}\rangle$\\
	\end{tabular}
};
\node (joinrb) [stdnodenum,left=-0.45em of joinr] {%
	\begin{tabular}{c}
		\multirow{1}{*}{$i$}\\
		\hline\hline
		$1$ \\
	\end{tabular}%
};
\node (joinl) [stdnode,left=1.6em of r12, yshift=-0.5em, label={[tdlabel, xshift=5.4em,yshift=-1em]above left:$\iota_{{6}}$}] {%
	\begin{tabular}{l@{\hspace{0.0em}}r}%
		\multirow{1}{*}{$\langle \tuplecolor{\outputPredColor}{\sigma_{6.i}}, $}&\multirow{1}{*}{$\tuplecolor{\statePredColor}{c_{6.i}} \rangle$}\\
		\hline\hline
		$\langle \tuplecolor{\outputPredColor}{\{\langle}\tuplecolor{\inputPredColor}{\{a\}}\tuplecolor{\outputPredColor}{\rangle\}}, $&$\tuplecolor{\statePredColor}{2}\rangle$\\
	\end{tabular}
};
\node (joinlb) [stdnodenum,left=-0.45em of joinl] {%
	\begin{tabular}{c}
		\multirow{1}{*}{$i$}\\
		\hline\hline
		$1$ \\
	\end{tabular}%
};
\coordinate (top) at ($ (leaf2.north east)+(0.6em,-0.5em) $);
\coordinate (bot) at ($ (top)+(0,-12.9em) $);

\draw [<-] (j) to (rt);
\draw [->] (j) to ($ (r12.north)$);
\draw [->] (j) to ($ (r2.north)$);
\draw [->](r2) to (i22);
\draw [<-](i2) to (i22);
\draw [<-](l2) to (i2);
\draw [<-](l1) to (i1);
\draw [->](i12) to (i1);
\draw [->](i13) to (i12);
\draw [->](r1) to (i13);
\draw [->](r12) to (r1);

\draw [dashed, bend left=22] (joinrrt) to (rt);
\draw [dashed] (j) to (join);
\draw [dashed, bend left=20] (i22) to (leaf2);
\draw [dashed, bend left=5] (i13) to (leaf1);
\draw [dashed, bend left=22] (leaf0) to (l1);
\draw [dashed, bend right=14] (leaf0x) to (r1);
\draw [dashed, bend right=18] (joinr) to (r2);
\draw [dashed, bend right=15] (joinl) to (r12);
\end{tikzpicture}
\caption{Selected tables obtained by~$\dpa_{\algo{PROJ}}$ on
  TD~${\cal T}'$ using~$\dpa_{\PRIM}$ (c.f.,
  Figure~\ref{fig:running1}).}
\label{fig:running2}
\end{figure}

\vspace{-0.75em}

\section{Runtime (Upper and Lower Bounds)}\label{sec:complexityresults}
In this section, we first present asymptotic upper bounds on the
runtime of our Algorithm~$\dpa_{\PROJ}$.  For the analysis, we
assume~$\gamma(n)$ to be the costs for multiplying two $n\hy$bit
integers, which can be achieved in time
$n\cdot log\, n \cdot log\, log\,
n$~\cite{Knuth1998,Harvey2016}. \longversion{Recently, an even faster
  algorithm was published~\cite{Harvey2016}.}

Then, our main result is a
lower bound that establishes that there cannot be an algorithm that
solves \PMC in time that is only single exponential in the treewidth
and polynomial in the size of the formula unless the exponential time
hypothesis (ETH) fails.
This result establishes that there \emph{cannot} be an algorithm
exploiting treewidth that is \emph{asymptotically better} than our
presented algorithm, although one can likely improve on the analysis
and give a better algorithm.
\longversion{One could for example cache~$\pcnt$ values, which, however, overcomplicates worst-case analysis.}
%



\begin{theorem}
  \label{thm:runtime}
  Given a \PMC instance~$(F,P)$ and a tree
  decomposition~${\cal T} = (T,\chi)$ of~$F$ of width~$k$ with $g$
  nodes. Algorithm~$\dpa_{\PROJ}$ runs in time
  $\mathcal{O}(2^{2^{k+4}}\cdot \gamma(\CCard{F}) \cdot g)$.
\end{theorem}
\begin{proof}
  Let~$d = k+1$ be maximum bag size of~$\TTT$. For each node~$t$ of $T$, we consider
  table $\tau=\ATab{\AlgS}[t]$ which has been computed by
  $\dpa_\AlgS$~\cite{SamerSzeider10b}. The table~$\tau$ has at most
  $2^{d}$ rows.
  In the worst case we store in~$\iota = \ATab{\PROJ}[t]$ each
  subset~$\sigma \subseteq \tau$ together with exactly one
  counter. Hence, we have $2^{2^{d}}$ many rows in $\iota$.
  In order to compute $\ipmc$ for~$\sigma$, we consider every
  subset~$\rho \subseteq \sigma$ and compute~$\pcnt$. Since
  $\Card{\sigma}\leq 2^d$, we have at most~$2^{2^{d}}$ many subsets
  $\rho$ of $\sigma$. 
For computing $\pcnt$, there could be each subset of the origins of~$\rho$ for each child
  table, which are less than~$2^{2^{d+1}}\cdot 2^{2^{d+1}}$ 
  (join and remove case).
  %
  %
  %
  In total, we obtain a runtime bound of~
  $\bigO{2^{2^{d}} \cdot 2^{2^{d}} \cdot 2^{2^{d+1}}\cdot 2^{2^{d+1}}
    \cdot  \gamma(\CCard{F}) } \subseteq \bigO{2^{2^{d+3}} \cdot \gamma(\CCard{F}) }$
  since we also need multiplication of counters.
  Then, we apply this to every node~$t$ of the tree decomposition,
  which results in running
  time~$\bigO{2^{2^{d+3}} \cdot \gamma(\CCard{F})  \cdot g}$.
  %
\end{proof}

\begin{corollary}\label{cor:runtime}
  Given an instance $(F,P)$ of \PMC where $F$ has
  treewidth~$k$. Algorithm~$\mdpa{\AlgS}$ runs in
  time~$\mathcal{O}(2^{2^{k+4}}\cdot \gamma(\CCard{F})  \cdot\CCard{F})$.
\end{corollary}
\begin{proof}
  We compute in time~$2^{\mathcal{O}(k^3)}\cdot\Card{V}$ a tree
  decomposition~${\cal T'}$ of width at most~$k$~\cite{Bodlaender96}
  of primal graph~$P_F$. Then, we run a decision version of the
  algorithm~$\dpa_{\AlgS}$ by Samer and Szeider~\cite{SamerSzeider10b}
  in time~$\mathcal{O}(2^k \cdot \gamma(\CCard{F}) \cdot
  \CCard{F})$. 
  Then, we again traverse the decomposition, thereby keeping
  rows that have a satisfying extension (``purging''), in
  time~$\mathcal{O}(2^k \cdot \CCard{F})$. 
  Finally, we run $\dpa_{\PROJ}$ and obtain the claim by
  Theorem~\ref{thm:runtime} and since~${\cal T'}$ has linearly many nodes~\cite{Bodlaender96}. 
\end{proof}

The next results also establish the lower bounds for our worst-cases.

\begin{theorem}
  Unless ETH fails, $\PMC$ cannot be solved in
  time~$2^{2^{o(k)}}\cdot \CCard{F}^{o(k)}$ for a given instance
  $(F,P)$ where~$k$ is the treewidth of the primal graph of~$F$.
\end{theorem}
\begin{proof}
  Assume for proof by contradiction that there is such an algorithm.
  %
  %
  We show that this contradicts a \longversion{very }recent
  result~\cite{LampisMitsou17}, which states that one cannot decide
  the validity of a
  \shortversion{QBF~\cite{BiereHeuleMaarenWalsh09,KleineBuningLettman99}}\longversion{quantified
    Boolean formula\footnote{For quantified Boolean formulas we refer
      to standard
      texts~\cite{BiereHeuleMaarenWalsh09,KleineBuningLettman99}.}}~$Q=\exists
  V_1. \forall V_2. E$ in time~$2^{2^{o(k)}}\cdot \CCard{E}^{o(k)}$
  under ETH. Given an instance~$(Q,k)$ of~$\exists\forall$-\SAT when
  parameterized by the treewidth~$k$ of~$E$
, we provide a reduction to an 
  instance~$(\forall{V_1}.\exists V_2.E',k)$ of~$\forall\exists$-\SAT
  where~$E'\equiv \neg E$ and $E'$ is in
  \shortversion{CNF}\longversion{conjunctive normal form}. Observe
  that the primal graphs of~$E$ and $E'$ are isomorphic and therefore
  have the same treewidth~$k$~\cite{LampisMitsou17}. Then,
  given an instance~$(\forall{V_1}.\exists V_2.E',k)$
  of~$\forall\exists$-\SAT when parameterized by the treewidth~$k$, we
  provide a reduction to an instance~$((F,P,n),k)$ 
  of decision version~$\PMC$-exactly-$n$
  of~$\PMC$ 
  such that $F=E$, $P=V_1$, and the number $n$ of
  solutions is exactly~$2^{\Card{V_1}}$.
  \longversion{It is easy to see that the}\shortversion{The}  reduction
  gives a yes instance~$((F,P,n),k)$ of~$\PMC$-exactly-$n$ 
  if and only if $(\forall{V_1}.\exists V_2.E',k)$ is a yes instance of~$\forall\exists$-\SAT. 
  The reduction is \longversion{in fact }also an fpt-reduction, since
  the treewidth of $F$ is exactly~$k$.
  %
  %
\end{proof}

\begin{corollary}
  Given an instance $(F,P)$ of \PMC where $F$ has treewidth~$k$. Then,
  Algorithm~$\mdpa{\AlgS}$ runs in
  time~$2^{2^{\Theta(k)}} \cdot \gamma(\CCard{F})\cdot \CCard{F} $.
\end{corollary}

\section{Correctness of the Algorithm}

In the following, we state definitions required for the correctness
proofs of our algorithm \PROJ. In the end, we only store rows that
are restricted to the bag content to maintain runtime bounds. 
\longversion{In
related work~\cite{SamerSzeider10b}, it was shown that this suffices
for table algorithm~$\AlgS$, i.e., \PRIM 
is both sound and
complete.} Similar to related work~\cite{FichteEtAl17a,SamerSzeider10b}, we proceed in two steps. First, we define properties of
so-called \emph{$\PROJ$-solutions up to~$t$}, and then restrict
these to~\emph{$\PROJ$-row solutions} at~$t$.

For the following statements, we assume that we have given an arbitrary instance~$(F,P)$ of \PMC and
a tree decomposition~$\TTT = (T,\chi)$ of
formula~$F$, where $T=(N, A,n)$, node $n \in N$ is the root and $\TTT$ is of width~$k$.
Moreover, for every~$t \in N$ of tree decomposition~$\TTT$, we let
$\ATabs{\AlgS}{t}$ be the tables that have been computed by running
algorithm~$\dpa_\AlgS$ for the dedicated input. Analogously, let
$\ATabs{\PROJ}{t}$ be the tables computed by running~$\dpa_\PROJ$ for
the input.

%
%

\begin{definition}\label{def:globalsol}
  Let~$\emptyset \subsetneq \sigma \subseteq \ATab{\AlgS}[t]$ be a
  table with $\sigma \in \subbuckets_P(\ATab{\AlgS}[t])$.
  \hspace{-0.2em}We define a \emph{${\PROJ}$-solution up to~$t$} to be the sequence
  $\langle \hat \sigma\rangle \hspace{-0.15em}=\hspace{-0.15em} \langle\PExt_{\leq t}(\sigma)\rangle$.
\end{definition}

%
%

Next, we recall that we can reconstruct all models from the tables.

\begin{proposition}\label{prop:sat}
  $I(\PExt_{\leq n}(\ATab{\AlgS}[n])) \hspace{-0.1em}=\hspace{-0.1em} I(\Exts) \hspace{-0.1em}=\hspace{-0.1em} \{J \in
    \ta{\var(F)} | J\models F\}.$
\end{proposition}
\begin{proof}[Idea]
\shortversion{We use a construction similar to Samer and Szeider~\cite{SamerSzeider10b} and Pichler, R\"ummele, and
  Woltran~\cite[Fig.~1]{PichlerRuemmeleWoltran10}, where we simply collect preceding rows.}
  \longversion{In fact, we can use the construction by Samer and
  Szeider~\cite{SamerSzeider10b} of the tables and extend them
  globally. Then, the extensions simply collect the corresponding,
  preceding rows. By taking the interpretation parts $I(\cdots)$ of
  these collected rows we obtain the set of all models of the formula.
  A similar construction is used by Pichler, R\"ummele, and
  Woltran~\cite[Fig.~1]{PichlerRuemmeleWoltran10}, however, hidden in
  an algorithm to enumerate solutions.}
\end{proof}

Before we present equivalence results between~$\ipmc_{\leq t}(\ldots)$
and the recursive version~$\ipmc(t, \ldots)$
(Definition~\ref{def:ipmc}) used during the computation of
$\dpa_\PROJ$, recall that~$\ipmc_{\leq t}$ and~$\pmc_{\leq t}$
(Definition~\ref{def:pmc}) are key to compute the projected model
count. The following corollary states that computing $\ipmc_{\leq n}$
at the root~$n$ actually suffices to compute the
projected model count~$\pmc_{\leq n}$ of the formula.

\begin{corollary}\label{cor:psat}
    $\ipmc_{\leq n}(\ATab{\AlgS}[n]) = \pmc_{\leq n}(\ATab{\AlgS}[n])
    =$\\ $\Card{I_P(\PExt_{\leq n}(\ATab{\AlgS}[n]))}$
    $=\hspace{-0.15em} \Card{I_P(\Exts)} = \Card{\{J \cap P
       \mid J \in \ta{\var(F)}, J\models F\}}$
\end{corollary}
\begin{proof}
  The corollary immediately follows from Proposition~\ref{prop:sat}
  and the observation that $\Card{\ATab{\AlgS}[n]} \leq 1$ by properties of algorithm~$\AlgS$ and
  since $\chi(n) = \emptyset$.
\end{proof}

The following lemma establishes that the \PROJ-solutions up to
root~$n$ of a given tree decomposition solve the \PMC problem.

\begin{lemma}\label{lem:global}
  The
  value~$c = \sum_{\langle\hat\sigma\rangle\text{ is a \PROJ-solution
      up to } n}\Card{I_P(\hat \sigma)}$ if and only if $c$ is the
  projected model count of~$F$ with respect to the set~$P$ of
  projection variables.
\end{lemma}
\begin{proof}
  Assume
  that~$c = \sum_{\langle\hat\sigma\rangle\text{ is a \PROJ-solution
      up to } n}\Card{I_P(\hat \sigma)}$. Observe that there can be at
  most one projected solution up to~$n$, since~$\chi(n)=\emptyset$. %
  If~$c=0$, then $\ATab{\AlgS}[n]$ contains no rows. Hence, $F$ has no
  models,~c.f., Proposition~\ref{prop:sat}, and obviously also no
  models projected to~$P$. Consequently, $c$ is the projected model
  count of~$F$.  
  If~$c>0$ we have by Corollary~\ref{cor:psat} that~$c$ is
  equivalent to the projected model count of~$F$ with respect to~$P$.
%
  %
  We proceed similar in the if direction. 
\end{proof}

In the following, we provide for a given node~$t$ and a given \PROJ-solution up to~$t$, the definition of a \PROJ-row solution at~$t$.

\begin{definition}\label{def:loctab}
Let~$t, t'\in N$ be nodes of a given tree decomposition~${\cal T}$, and $\hat\sigma$ be a~\PROJ-solution up to~$t$. Then, we define \emph{the local table for}~$t'$ as
  $\local(t',\hat\sigma)\eqdef \{ \langle \vec{\tabval}\rangle |$ $
  \langle t', \vec{\tabval}\rangle \in \hat\sigma\}$, and
 if~$t=t'$, the \emph{$\PROJ$-row solution at $t$} by
  $\langle \local(t,\hat\sigma), \Card{I_P(\hat\sigma)}\rangle$.
  %
  %

\end{definition}






\begin{observation}\label{obs:unique}
  Let $\langle \hat \sigma\rangle$ be a \PROJ-solution up to a
  node~$t\in N$.  There is exactly one corresponding \PROJ-row
  solution
  $\langle \local(t,\hat\sigma), \Card{I_P(\hat\sigma)}\rangle$ at~$t$.

  Vice versa, let $\langle \sigma, c\rangle$ be a \PROJ-row
  solution at~$t$ for some integer~$c$. Then, there is exactly one
  corresponding \PROJ-solution~$\langle\PExt_{\leq t}(\sigma)\rangle$
  up to~$t$.
\end{observation}

We need to ensure that storing~$\PROJ$-row solutions at a
node suffices to solve the~\PMC problem, which is necessary
to obtain runtime bounds (c.f. Corollary~\ref{cor:runtime}).

\begin{lemma}\label{lem:local}
  Let $t\in N$ be a node of the tree decomposition~$\TTT$.  There is a
  \PROJ-row solution at root~$n$ if and only if the projected
  model count of~$F$ is larger than~$0$.
\end{lemma}
\begin{proof}%

  (``$\Longrightarrow$''): Let $\langle \sigma, c\rangle$ be a
  \PROJ-row solution at root~$n$ where $\sigma$ is a $\AlgS$-table and
  $c$ is a positive integer. Then, by Definition~\ref{def:loctab},
  there also exists a
  corresponding~$\PROJ$-solution~$\langle \hat\sigma \rangle$ up
  to~$n$ such that $\sigma = \local(n,\hat\sigma)$ and
  $c=\Card{I_P(\hat\sigma)}$.
  Moreover, since~$\chi(n)=\emptyset$, we
  have~$\Card{\ATab{\AlgS}[n]}=1$.  
  Then, by Definition~\ref{def:globalsol},
  $\hat \sigma = \ATab{\AlgS}[n]$. By Corollary~\ref{cor:psat}, we
  have $c=\Card{I_P(\ATab{\AlgS}[n])}$.
  Finally, the claim follows.
  %
%
  \quad(``$\Longleftarrow$''): Similar to the only-if
  direction.
\end{proof}

\begin{observation}\label{obs:main_incl_excl}
  Let $X_1$, $\ldots$, $X_n$ be  finite sets. 
  The number~$\Card{\bigcap_{i \in X} X_i}$ is given by
  $\Card{\bigcap_{i \in X} X_i} = \big|\Card{\bigcup^n_{j = 1} X_j} + \sum_{\emptyset \subsetneq I \subsetneq X} (-1)^{\Card{I}} 
                                              \Card{\bigcap_{i \in I} X_i}\big|.$
  \longversion{\begin{align*}
    \Card{\bigcap_{i \in X} X_i} 
    =& %
       \Bigg|\Card{\bigcup^n_{j = 1} X_j} &&- %
                                       \sum_{\emptyset \subsetneq I \subsetneq X, \Card{I}=1}
                                       \Card{\bigcap_{i \in I} X_i} + %
                                       \sum_{\emptyset \subsetneq I \subsetneq X, \Card{I}=2}
                                       \Card{\bigcap_{i \in I} X_i} - \ldots \\
                                         & &&+ \sum_{\emptyset \subsetneq I \subsetneq X, \Card{I}=n-1} (-1)^{\Card{I}} 
                                              \Card{\bigcap_{i \in I} X_i}\Bigg|.
  \end{align*}}
\end{observation}

\begin{lemma}\label{lem:main_incl_excl}
  Let $t\in N$ be a node of the tree decomposition~$\TTT$
  with~$\children(t,T) = \langle t_1, \ldots, t_\ell \rangle$ and let
  $\langle\sigma,\cdot\rangle$ be a~\PROJ-row solution at~$t$.
  Then, \vspace{-0.75em}
  \begin{enumerate}
  \item %
    $\ipmc(t,\sigma,\langle\ATab{\PROJ}[t_1],\makebox[1em][c]{.\hfil.\hfil.},
    \ATab{\PROJ}[t_{\ell}]\rangle) \hspace{-0.15em}=\hspace{-0.15em} \ipmc_{\leq t}(\sigma)$
  \item %
    If $\type(t) \neq \leaf$: $\pmc(t,\sigma,\langle\ATab{\PROJ}[t_1],\makebox[1em][c]{.\hfil.\hfil.},
    \ATab{\PROJ}[t_{\ell}]\rangle) \hspace{-0.15em}=\hspace{-0.15em} \pmc_{\leq t}(\sigma)$.
  \end{enumerate}
\end{lemma}
\begin{proof}[Sketch]
  We prove the statement by simultaneous induction.
  (``Induction Hypothesis''): Lemma~\ref{lem:main_incl_excl} holds for the nodes in~$\children(t,T)$ and also for node~$t$, but on strict subsets~$\rho\subsetneq\sigma$.
  (``Base Cases''): Let $\type(t) = \leaf$.
  By definition,
  $\ipmc(t,\emptyset, \langle \rangle) = \ipmc_{\leq t}(\emptyset) =
  1$.
  Recall that for $\pmc$ the equivalence does not hold for leaves, but we use a node~$t$
  that has a node~$t'\in N$ with~$\type(t') = \leaf$ as child for the
  base case. Observe that by definition~$t$ has 
  exactly one child.
  Then, we have
  $\pmc(t,\sigma,\langle\ATab{\PROJ}[t']\rangle) = \sum_{\emptyset
    \subsetneq O \subseteq {\origs(t,\sigma)}} (-1)^{(\Card{O} - 1)}
  \cdot \sipmc(\langle \ATab{\AlgS}[t']\rangle, O) =
  \Card{\bigcup_{\vec u\in\sigma} I_P(\PExt_{\leq t}(\{\vec u\}))} =
  \pmc_{\leq t}(\sigma) = 1$ for \PROJ-row
  solution~$\langle\sigma,\cdot\rangle$ at~$t$.
  (``Induction Step''): We proceed by case distinction.
  Assume that $\type(t) = \intr$.
  Let $a \in (\chi(t) \setminus \chi(t'))$ be the introduced
  variable. We have two cases. Assume Case (i): $a$ also belongs to
  $(\var(F) \setminus P)$,~i.e., $a$ is not a projection variable. 
  %
  %
  %
  Let~$\langle \sigma, c \rangle$ be a \PROJ-row solution at~$t$ for
  some integer~$c$. By construction of algorithm~$\AlgS$
  there are many rows in the table~$\ATab{\AlgS}[t]$ for one row in
  the table~$\ATab{\AlgS}[t']$, more precisely,
  $\Card{\buckets_P(\sigma)} = 1$.
  As a result,
  $\pmc_{\leq t}(\sigma) = \pmc_{\leq t'}(\orig(t,\sigma))$ by
  applying Observation~\ref{obs:unique}.
  We apply the inclusion-exclusion principle on every subset~$\rho$ of
  the origins of~$\sigma$ in the definition of~$\pmc$ and by induction
  hypothesis we know that
  $\ipmc(t',\rho,\langle\ATab{\PROJ}[t']\rangle) = \ipmc_{\leq
    t'}(\rho)$, therefore,
  $\sipmc(\ATab{\PROJ}[t'], \rho) = \ipmc_{\leq t'}(\rho)$.  This
  concludes Case~(i) for $\pmc$. The induction step for $\ipmc$ works
  similar by applying
  Observation~\ref{obs:main_incl_excl} and comparing corresponding
  \PROJ-solutions up to~$t$ or $t'$, respectively. Further, for showing the lemma for~$\ipmc$, one has to additionally apply the hypothesis for node~$t$, but on strict subsets~$\emptyset\subsetneq\rho\subsetneq\sigma$ of~$\sigma$.
  %
  Assume that we have Case~(ii): $a$ also belongs to $P$,~i.e., $a$ is a projection
  variable. This is a special case of Case~(i) since
  $\Card{\buckets_P(\sigma)} = 1$. Similarly, for join and remove nodes. 
%
\end{proof}


\begin{lemma}[Soundness]\label{lem:correct}
  Let $t\in N$ be a node of the tree decomposition~$\TTT$ with
  $\children(t,T) = \langle t_1, \ldots, t_\ell \rangle$.
  Then, each row~$\langle \tab{}, c \rangle$ at node~$t$ obtained by~$\PROJ$ 
  is a~\PROJ-row solution for~$t$.
\end{lemma}
\begin{proof}[Idea]
  Observe that Listing~\ref{fig:dpontd3} computes a row for each
  sub-bucket $\sigma \in \subbuckets_P(\ATab{\AlgS}[t])$. The
  resulting row~$\langle\sigma, c \rangle$ obtained by~$\ipmc$ is
  indeed a \PROJ-row solution for~$t$ according to
  Lemma~\ref{lem:main_incl_excl}.
\end{proof}


\begin{lemma}[Completeness]\label{lem:complete}
  Let~$t\in N$ be a node of tree decomposition~$\TTT$ where~$\children(t,T) = \langle t_1, \ldots, t_\ell \rangle$
  and~$\type(t) \neq \leaf$. Given a 
	  \PROJ-row solution~$\langle \sigma, c \rangle$ at~$t$.
  Then, there is $\langle C_1, \ldots, C_\ell\rangle$ where each $C_i$ is a set
  of \PROJ-row solutions at~$t_i$
  with
  $\sigma = \PROJ(t, \cdot, \cdot, P, \langle C_1, \ldots,
  C_\ell\rangle, \ATab{\AlgS})$.
\end{lemma}
\begin{proof}[Idea]
Since~$\langle\sigma,c \rangle$ is a~\PROJ-row solution for~$t$, there is by Definition~\ref{def:loctab} a corresponding ~\PROJ-solution~$\langle\hat\sigma\rangle$ up to~$t$ such that~$\local(t,\hat\sigma) = \sigma$. 
We proceed again by case distinction. Assume~$\type(t)=\intr$ and~$t'=t_1$. Then we define~$\hat{\sigma'}\eqdef \{(t',\hat\rho) \mid (t', \hat\rho)\in \sigma, t \neq t'\}$. Then, for each subset~$\emptyset\subsetneq\rho\subseteq\local(t',\hat{\sigma'})$, we define~$\langle \rho, \Card{I_P(\PExt_{\leq t}(\rho))}\rangle$ in accordance with Definition~\ref{def:loctab}. By Observation~\ref{obs:unique}, we have that~$\langle \rho, \Card{I_P(\PExt_{\leq t}(\rho))}\rangle$ is a \AlgS-row solution at~$t'$. 
Since we defined~\PROJ-row solutions for~$t'$ for all respective \PROJ-solutions up to~$t'$, we encountered every~\PROJ-row solution for~$t'$ 
required for deriving~$\langle \sigma, c\rangle$ via~\PROJ (c.f. Definitions~\ref{def:ipmc} and~\ref{def:pcnt}).
%
%
Similarly, for remove and join nodes. 
\end{proof}


\begin{theorem}\label{thm:correctness}
  The algorithm~$\dpa_\PROJ$ is correct. 
  More precisely, 
  %
  $\dpa_\PROJ((F,P),\TTT,$ $\ATab{\AlgS})$ returns
  tables~$\ATab{\PROJ}$ such that $c=\sipmc(\ATab{\AlgS}[n], \cdot)$
  is the projected model count of~$F$ with respect to the set~$P$ of
  projection variables.
\end{theorem}
\begin{proof}
  %
  By Lemma~\ref{lem:correct} we have soundness for every
  node~$t \in N$ and hence only valid rows as output of table
  algorithm~$\PROJ$ when traversing the tree decomposition in
  post-order up to the root~$n$.
  By Lemma~\ref{lem:local} we know that the projected model count~$c$
  of~$F$ is larger than zero if and only if there exists a
  certain~\PROJ-row solution for~$n$.
  This~\PROJ-row solution at node~$n$ is of the
  form~$\langle \{\langle\emptyset, \ldots\rangle\} ,c\rangle$. If
  there is no \PROJ-row solution at node~$n$,
  then~$\ATab{\AlgS}[n]=\emptyset$ since the table algorithm~$\AlgS$
  is correct (c.f. Proposition~\ref{prop:sat}). Consequently, we have
  $c=0$. Therefore, $c=\sipmc(\ATab{\AlgS}[n], \cdot)$ is the
  pmc of~$F$ w.r.t.~$P$ in both cases.
  %
  %
  
  %
  %

  %
  %

  Next, we establish completeness by induction starting from 
  root~$n$. Let therefore, $\langle \hat\sigma \rangle$ be the~\PROJ-solution up to~$n$, where for each row
  in~$\vec u\in \hat\sigma$, $I(\vec u)$ corresponds to a model of~$F$.  By
  Definition~\ref{def:loctab}, we know that for~$n$ we
  can construct a \PROJ-row solution at~$n$ of the
  form~$\langle \{\langle\emptyset, \ldots\rangle\} ,c\rangle$
  for~$\hat\sigma$.  We already established the induction step in
  Lemma~\ref{lem:complete}.
  \longversion{Hence, we obtain some row for every
  node~$t$.}
  Finally, we stop at the leaves.
  %
%
%
\end{proof}

\begin{corollary}\label{cor:correctness}
  The algorithm $\mdpa{\AlgS}$ is correct, i.e., $\mdpa{\AlgS}$ solves~\PMC. 
\end{corollary}
\begin{proof}
  The result follows, since~$\mdpa{\AlgS}$ consists of
  pass~$\dpa_\AlgS$, a purging step and~$\dpa_\PROJ$. For
  correctness of~$\dpa_\AlgS$ we refer to other
  sources~\cite{FichteEtAl17a,SamerSzeider10b}. By Proposition~\ref{prop:sat}, ``purging'' neither destroys soundness nor completeness
  of~$\dpa_\PRIM$.
\end{proof}

\section{Conclusions}\label{sec:conclusions}
We introduced a dynamic programming algorithm to solve projected model
counting (\PMC) by exploiting the structural parameter treewidth. Our
algorithm is asymptotically optimal under the exponential time
hypothesis (ETH). Its runtime is double exponential in the treewidth
of the primal 
graph of the instance and polynomial in the size of the input
instance. We believe that our results can also be extended to another
graph representation, namely the incidence graph.
Our approach 
is very general and might be applicable to a wide range of other hard
combinatorial problems, such as projection for
ASP~\cite{FichteEtAl17a} and QBF~\cite{CharwatWoltran16a}.
%
%

\appendix
\newpage
\bibliography{references}

\begin{thebibliography}{10}

\bibitem{AbiteboulHullVianu95}
S.~Abiteboul, R.~Hull, and V.~Vianu.
\newblock {\em Foundations of Databases: The Logical Level}.
\newblock Addison-Wesley, Boston, MA, USA, 1st edition, 1995.

\bibitem{AbramsonBrownEdwards96a}
B.~Abramson, J.~Brown, W.~Edwards, A.~Murphy, and R.~L. Winkler.
\newblock Hailfinder: A {B}ayesian system for forecasting severe weather.
\newblock {\em International Journal of Forecasting}, 12(1):57--71, 1996.

\bibitem{AzizChuMuise15a}
R.~A. Aziz, G.~Chu, C.~Muise, and P.~Stuckey.
\newblock {\#($\exists$)SAT: Projected Model Counting}.
\newblock In M.~Heule and S.~Weaver, editors, {\em Proceedings of the 18th
  International Conference on Theory and Applications of Satisfiability Testing
  (SAT'15)}, pages 121--137, Austin, TX, USA, Sept. 2015. Springer Verlag.

\bibitem{BiereHeuleMaarenWalsh09}
A.~Biere, M.~Heule, H.~van Maaren, and T.~Walsh, editors.
\newblock {\em Handbook of Satisfiability}, volume 185 of {\em Frontiers in
  Artificial Intelligence and Applications}.
\newblock IOS Press, Amsterdam, Netherlands, Feb. 2009.

\bibitem{Bodlaender96}
H.~L. Bodlaender.
\newblock A linear-time algorithm for finding tree-decompositions of small
  treewidth.
\newblock {\em SIAM J. Comput.}, 25(6):1305--1317, 1996.

\bibitem{BodlaenderKloks96}
H.~L. Bodlaender and T.~Kloks.
\newblock Efficient and constructive algorithms for the pathwidth and treewidth
  of graphs.
\newblock {\em J. Algorithms}, 21(2):358--402, 1996.

\bibitem{BodlaenderKoster08}
H.~L. Bodlaender and A.~M. C.~A. Koster.
\newblock Combinatorial optimization on graphs of bounded treewidth.
\newblock {\em The Computer Journal}, 51(3):255--269, 2008.

\bibitem{BondyMurty08}
J.~A. Bondy and U.~S.~R. Murty.
\newblock {\em Graph theory}, volume 244 of {\em Graduate Texts in
  Mathematics}.
\newblock Springer Verlag, New York, USA, 2008.

\bibitem{ChakrabortyMeelVardi16a}
S.~Chakraborty, K.~S. Meel, and M.~Y. Vardi.
\newblock Improving approximate counting for probabilistic inference: From
  linear to logarithmic {SAT} solver calls.
\newblock In S.~Kambhampati, editor, {\em Proceedings of 25th International
  Joint Conference on Artificial Intelligence (IJCAI'16)}, pages 3569--3576,
  New York City, NY, USA, July 2016. The AAAI Press.

\bibitem{CharwatWoltran16a}
G.~Charwat and S.~Woltran.
\newblock Dynamic programming-based {QBF} solving.
\newblock In F.~Lonsing and M.~Seidl, editors, {\em Proceedings of the 4th
  International Workshop on Quantified Boolean Formulas (QBF'16)}, volume 1719,
  pages 27--40. CEUR Workshop Proceedings (CEUR-WS.org), 2016.
\newblock co-located with 19th International Conference on Theory and
  Applications of Satisfiability Testing (SAT'16).

\bibitem{ChoiBroeckDarwiche15a}
A.~Choi, G.~Van~den Broeck, and A.~Darwiche.
\newblock Tractable learning for structured probability spaces: A case study in
  learning preference distributions.
\newblock In Q.~Yang, editor, {\em Proceedings of 24th International Joint
  Conference on Artificial Intelligence (IJCAI'15)}. The AAAI Press, 2015.

\bibitem{CyganEtAl15}
M.~Cygan, F.~V. Fomin, {\L}.~Kowalik, D.~Lokshtanov, M.~P. D{\'a}niel~Marx,
  M.~Pilipczuk, and S.~Saurabh.
\newblock {\em Parameterized Algorithms}.
\newblock Springer Verlag, 2015.

\bibitem{Diestel12}
R.~Diestel.
\newblock {\em Graph Theory, 4th Edition}, volume 173 of {\em Graduate Texts in
  Mathematics}.
\newblock Springer Verlag, 2012.

\bibitem{DomshlakHoffmann07a}
C.~Domshlak and J.~Hoffmann.
\newblock Probabilistic planning via heuristic forward search and weighted
  model counting.
\newblock {\em J. Artif. Intell. Res.}, 30:565--620, 2007.

\bibitem{DowneyFellows13}
R.~G. Downey and M.~R. Fellows.
\newblock {\em Fundamentals of Parameterized Complexity}.
\newblock Texts in Computer Science. Springer Verlag, London, UK, 2013.

\bibitem{MeelEtAl17a}
L.~Due{\~{n}}as{-}Osorio, K.~S. Meel, R.~Paredes, and M.~Y. Vardi.
\newblock Counting-based reliability estimation for power-transmission grids.
\newblock In S.~P. Singh and S.~Markovitch, editors, {\em Proceedings of the
  Thirty-First {AAAI} Conference on Artificial Intelligence (AAAI'17)}, pages
  4488--4494, San Francisco, CA, {USA}, Feb. 2017. The AAAI Press.

\bibitem{DurandHermannKolaitis05}
A.~Durand, M.~Hermann, and P.~G. Kolaitis.
\newblock Subtractive reductions and complete problems for counting complexity
  classes.
\newblock {\em Theoretical Computer Science}, 340(3):496--513, 2005.

\bibitem{FichteEtAl17a}
J.~K. Fichte, M.~Hecher, M.~Morak, and S.~Woltran.
\newblock Answer set solving with bounded treewidth revisited.
\newblock In M.~Balduccini and T.~Janhunen, editors, {\em Proceedings of the
  14th International Conference on Logic Programming and Nonmonotonic Reasoning
  (LPNMR'17)}, volume 10377 of {\em Lecture Notes in Computer Science}, pages
  132--145, Espoo, Finland, July 2017. Springer Verlag.

\bibitem{FichteEtAl17b}
J.~K. Fichte, M.~Hecher, M.~Morak, and S.~Woltran.
\newblock {DynASP2.5}: Dynamic programming on tree decompositions in action.
\newblock In D.~Lokshtanov and N.~Nishimura, editors, {\em Proceedings of the
  12th International Symposium on Parameterized and Exact Computation
  (IPEC'17)}. Dagstuhl Publishing, 2017.

\bibitem{FlumGrohe06}
J.~Flum and M.~Grohe.
\newblock {\em Parameterized Complexity Theory}, volume XIV of {\em Theoretical
  Computer Science}.
\newblock Springer Verlag, Berlin, 2006.

\bibitem{GebserKaufmannSchaub09a}
M.~Gebser, B.~Kaufmann, and T.~Schaub.
\newblock Solution enumeration for projected boolean search problems.
\newblock In W.-J. van Hoeve and J.~N. Hooker, editors, {\em Proceedings of the
  6th International Conference on Integration of AI and OR Techniques in
  Constraint Programming for Combinatorial Optimization Problems (CPAIOR'09)},
  volume 5547 of {\em Lecture Notes in Computer Science}, pages 71--86, Berlin,
  2009. Springer Verlag.

\bibitem{GebserSchaubThieleVeber11}
M.~Gebser, T.~Schaub, S.~Thiele, and P.~Veber.
\newblock Detecting inconsistencies in large biological networks with answer
  set programming.
\newblock {\em Theory Pract. Log. Program.}, 11(2-3):323--360, 2011.

\bibitem{GinsbergParkesRoy98a}
M.~L. Ginsberg, A.~J. Parkes, and A.~Roy.
\newblock Supermodels and robustness.
\newblock In C.~Rich and J.~Mostow, editors, {\em Proceedings of the 15th
  National Conference on Artificial Intelligence and 10th Innovative
  Applications of Artificial Intelligence Conference (AAAI/IAAI'98)}, pages
  334--339, Madison, Wisconsin, USA, July 1998. The AAAI Press.

\bibitem{GomesKautzSabharwalSelman08a}
C.~P. Gomes, A.~Sabharwal, and B.~Selman.
\newblock Chapter 20: Model counting.
\newblock In A.~Biere, M.~Heule, H.~van Maaren, and T.~Walsh, editors, {\em
  Handbook of Satisfiability}, volume 185 of {\em Frontiers in Artificial
  Intelligence and Applications}, pages 633--654. IOS Press, Amsterdam,
  Netherlands, Feb. 2009.

\bibitem{GrahamGrotschelLovasz95a}
R.~L. Graham, M.~Gr{\"o}tschel, and L.~Lov{\'a}sz.
\newblock {\em Handbook of Combinatorics}, volume~I.
\newblock Elsevier Science Publishers, North-Holland, 1995.

\bibitem{Harvey2016}
D.~Harvey, J.~van~der Hoeven, and G.~Lecerf.
\newblock Even faster integer multiplication.
\newblock {\em J. Complexity}, 36:1--30, 2016.

\bibitem{HemaspaandraVollmer95a}
L.~A. Hemaspaandra and H.~Vollmer.
\newblock The satanic notations: Counting classes beyond \#{P} and other
  definitional adventures.
\newblock {\em SIGACT News}, 26(1):2--13, Mar. 1995.

\bibitem{ImpagliazzoPaturiZane01}
R.~Impagliazzo, R.~Paturi, and F.~Zane.
\newblock Which problems have strongly exponential complexity?
\newblock {\em J. of Computer and System Sciences}, 63(4):512--530, 2001.

\bibitem{KleineBuningLettman99}
H.~Kleine~B{\"u}ning and T.~Lettman.
\newblock {\em Propositional logic: deduction and algorithms}.
\newblock Cambridge University Press, Cambridge, New York, NY, USA, 1999.

\bibitem{Knuth1998}
D.~E. Knuth.
\newblock How fast can we multiply?
\newblock In {\em The Art of Computer Programming}, volume~2 of {\em
  Seminumerical Algorithms}, chapter 4.3.3, pages 294--318. Addison-Wesley, 3
  edition, 1998.

\bibitem{LagniezMarquis17a}
J.-M. Lagniez and P.~Marquis.
\newblock An improved decision-{DNNF} compiler.
\newblock In C.~Sierra, editor, {\em Proceedings of the Twenty-Sixth
  International Joint Conference on Artificial Intelligence (IJCAI'17)}. The
  AAAI Press, 2017.

\bibitem{LampisMitsou17}
M.~Lampis and V.~Mitsou.
\newblock Treewidth with a quantifier alternation revisited.
\newblock In D.~Lokshtanov and N.~Nishimura, editors, {\em Proceedings of the
  12th International Symposium on Parameterized and Exact Computation
  (IPEC'17)}. Dagstuhl Publishing, 2017.

\bibitem{ManningRaghavanSchutze08a}
C.~D. Manning, P.~Raghavan, and H.~Sch{\"u}tze.
\newblock {\em Introduction to Information Retrieval}.
\newblock Cambridge University Press, Cambridge, 2008.

\bibitem{Niedermeier06}
R.~Niedermeier.
\newblock {\em Invitation to Fixed-Parameter Algorithms}, volume~31 of {\em
  Oxford Lecture Series in Mathematics and its Applications}.
\newblock Oxford University Press, New York, NY, USA, 2006.

\bibitem{Papadimitriou94}
C.~H. Papadimitriou.
\newblock {\em Computational Complexity}.
\newblock Addison-Wesley, 1994.

\bibitem{PichlerRuemmeleWoltran10}
R.~Pichler, S.~R{\"u}mmele, and S.~Woltran.
\newblock Counting and enumeration problems with bounded treewidth.
\newblock In E.~M. Clarke and A.~Voronkov, editors, {\em Proceedings of the
  16th International Conference on Logic for Programming, Artificial
  Intelligence, and Reasoning (LPAR'10)}, volume 6355 of {\em Lecture Notes in
  Computer Science}, pages 387--404. Springer Verlag, 2010.

\bibitem{PourretNaimBruce08a}
O.~Pourret, P.~Naim, and M.~Bruce.
\newblock {\em Bayesian Networks - A Practical Guide to Applications}.
\newblock John Wiley \& Sons, 2008.

\bibitem{Roth96a}
D.~Roth.
\newblock On the hardness of approximate reasoning.
\newblock {\em Artificial Intelligence}, 82(1--2), 1996.

\bibitem{SaetherTelleVatshelle15a}
S.~H. S{\ae}ther, J.~A. Telle, and M.~Vatshelle.
\newblock Solving \#{SAT} and {MAXSAT} by dynamic programming.
\newblock {\em J. Artif. Intell. Res.}, 54:59--82, 2015.

\bibitem{SahamiDumaisHeckerman98a}
M.~Sahami, S.~Dumais, D.~Heckerman, and E.~Horvitz.
\newblock A {B}ayesian approach to filtering junk e-mail.
\newblock In T.~Joachims, editor, {\em Proceedings of the AAAI-98 Workshop on
  Learning for Text Categorization}, volume~62, pages 98--105, 1998.

\bibitem{SamerSzeider10b}
M.~Samer and S.~Szeider.
\newblock Algorithms for propositional model counting.
\newblock {\em J. Discrete Algorithms}, 8(1):50---64, 2010.

\bibitem{SangBeameKautz05a}
T.~Sang, P.~Beame, and H.~Kautz.
\newblock Performing {B}ayesian inference by weighted model counting.
\newblock In M.~M. Veloso and S.~Kambhampati, editors, {\em Proceedings of the
  29th National Conference on Artificial Intelligence (AAAI'05)}. The AAAI
  Press, 2005.

\bibitem{Valiant79}
L.~Valiant.
\newblock The complexity of enumeration and reliability problems.
\newblock {\em SIAM J. Comput.}, 8(3):410--421, 1979.

\bibitem{Wilder12a}
R.~L. Wilder.
\newblock {\em Introduction to the Foundations of Mathematics}.
\newblock John Wiley \& Sons, 2nd edition edition, 1965.

\bibitem{XueChoiDarwiche12a}
Y.~Xue, A.~Choi, and A.~Darwiche.
\newblock Basing decisions on sentences in decision diagrams.
\newblock In J.~Hoffmann and B.~Selman, editors, {\em Proceedings of the 26th
  AAAI Conference on Artificial Intelligence (AAAI'12)}, Toronto, ON, Canada,
  2012. The AAAI Press.

\end{thebibliography}


\longversion{
\appendix
\newpage

\section{Omitted Definitions}
A \emph{(prenex) quantified Boolean formula}~$\Q$ is of the
form
  $Q_1 V_1. Q_2 V_2.\ldots Q_m V_m. F$
  where $Q_i \in \{\forall, \exists\}$, $V_i$ are disjoint sets of
  Boolean variables, and $F$ is a Boolean formula that contains only
  the variables in $\bigcup^m_{i=1} V_i$.
  %
  %
  The truth (evaluation) of quantified Boolean formulas is defined in the standard way.
  Given a quantified Boolean formula~$\Q$, the evaluation problem of
  quantified Boolean formulas~\QBFSAT asks whether $Q$ evaluates to
  true.
  The problem~\QBFSAT is \PSPACE-complete and is therefore believed to
  be computationally harder than
  \SAT~\cite{KleineBuningLettman99,Papadimitriou94,StockmeyerMeyer73}.
  A well known fragment of \QBFSAT is $\forall\exists$\hy \SAT where
  the input is restricted to quantified Boolean formulas of the
  form~$\forall V_1.\exists V_2.F$ where $F$ is a propositional
  CNF formula. The complexity class consisting of all problems that are
  polynomial-time reducible to $\forall\exists$\hy \SAT is denoted by
  $\Pi_2^P$, and its complement is denoted by $\Sigma_2^P$. 
For more
detailed information on QBFs we refer to other sources,
e.g.,~\cite{BiereHeuleMaarenWalsh09,KleineBuningLettman99}.



\section{Omitted Proofs}

\begin{restateobservation}[obs:relation]%
\begin{observation}
The relation~$\bucket$ is an equivalence relation.
\end{observation}
\end{restateobservation}
\begin{proof}
One can easily see that~$=_P(A,B) \eqdef (A \cap P) = (B \cap P)$ is 
\begin{itemize}
	\item reflexive: $A\cap P = A \cap P$ for any two sets~$A, P$,
	\item symmetric: $A\cap P = B \cap P$ if and only if
          $B\cap P = A\cap P$ for given sets~$A,B,P$, and
	\item transitive: if $A \cap P = B \cap P$  and~$B \cap P = C \cap P$, then~$A\cap P = C\cap P$ holds as well for any sets~$A,B,C,P$.
\end{itemize}
As a result, $=_P$ is an equivalence relation.
\end{proof}

\paragraph{\textbf{Global Assumptions.}}
Here, we fix requirements for almost all statements in the following.
We assume that we have given an arbitrary instance~$(F,P)$ of \PMC and
a tree decomposition~$\TTT = (T,\chi)$ of
formula~$F$, where $T=(N, A,n)$, node $n \in N$ is the root and $\TTT$ is of width~$k$.
Moreover, for every~$t \in N$ of tree decomposition~$\TTT$, we let
$\ATabs{\AlgS}{t}$ be the tables that have been computed by running
algorithm~$\dpa_\AlgS$ for the dedicated input. Analogously, let
$\ATabs{\PROJ}{t}$ be the tables that have been computed by running
algorithm~$\dpa_\PROJ$ for the dedicated input.

\begin{lemma}\label{lem:runtime}
  For every node~$t\in N$, $\ATabs{\PROJ}{t}$ contains at
  most~$2^{2^{k+1}}$ rows.
\end{lemma}
\begin{proof}
  By definition of a tree decomposition, for every node~$t\in N$ the
  bag~$\chi(t)$ is of size at most~$k+1$. Therefore, we have at
  most~$2^{k+1}$ many~\cite{SamerSzeider10b} rows in the table
  obtained via~$\dpa_\AlgS$. In the end, we have at most $2^{2^{k+1}}$
  rows within table~$\ATabs{\PROJ}{t}$, since we have a row for each
  of the $2^{2^{k+1}}$ many subsets of a $\AlgS$-row.
\end{proof}

In the following, we state definitions required for the correctness
proofs of our algorithm \PROJ.  In the end, we only store rows that
are restricted to the bag content to maintain runtime bounds. In
related work~\cite{SamerSzeider10b}, it was shown that this suffices
for table algorithm~$\AlgS$, i.e., \PRIM and~\INC are both sound and
complete.  Similar to related work~\cite{FichteEtAl17a}, we define the
content of our tables in two steps. First, we define the properties of
so-called \emph{$\PROJ$-solutions up to~$t$}. Second, we restrict
these solutions to~\emph{$\PROJ$-row solutions} at~$t$.

%
%

\begin{definition}\label{def:globalsol}
  Let~$\emptyset \subsetneq \sigma \subseteq \ATab{\AlgS}[t]$ be a
  table with $\sigma \in \subbuckets_P(\ATab{\AlgS}[t])$.
  We define a \emph{${\PROJ}$-solution up to~$t$} to be the sequence
  $\langle \hat \sigma\rangle = \langle\PExt_{\leq t}(\sigma)\rangle$.
\end{definition}

%
%

Next, we recall that we can reconstruct all models from the tables.

\begin{proposition}[c.f.~\cite{SamerSzeider10b}]\label{prop:sat}
  \[I(\PExt_{\leq n}(\ATab{\AlgS}[n])) = I(\Exts) = \{J \mid J \in
    \ta{\var(F)}, J\models F\}.\]
\end{proposition}
\begin{proof}[Idea]
  In fact, we can use the construction by Samer and
  Szeider~\cite{SamerSzeider10b} of the tables and extend them
  globally. Then, the extensions simply collect the corresponding,
  preceding rows. By taking the interpretation parts $I(\cdots)$ of
  these collected rows we obtain the set of all models of the formula.
  A similar construction is used by Pichler, R\"ummele, and
  Woltran~\cite[Fig.~1]{PichlerRuemmeleWoltran10}, however, hidden in
  an algorithm to enumerate solutions.
\end{proof}

Before we present equivalence results between~$\ipmc_{\leq t}(\ldots)$
and the recursive version~$\ipmc(t, \ldots)$
(Definition~\ref{def:ipmc}) used during the computation of
$\dpa_\PROJ$, recall that~$\ipmc_{\leq t}$ and~$\pmc_{\leq t}$
(Definition~\ref{def:pmc}) are key to compute the projected model
count. The following corollary states that computing $\ipmc_{\leq n}$
at the root~$n$ actually suffices to compute~$\pmc_{\leq n}$, which is
in fact the projected model count of the input formula.

\begin{corollary}\label{cor:psat}
  \begin{align*}
    \ipmc_{\leq n}(\ATab{\AlgS}[n]) =& \pmc_{\leq n}(\ATab{\AlgS}[n])\\
    =& \Card{I_P(\PExt_{\leq n}(\ATab{\AlgS}[n]))}\\
    =& \Card{I_P(\Exts)}\\
    =& \Card{\{J \cap P
       \mid J \in \ta{\var(F)}, J\models F\}}
  \end{align*}
\end{corollary}
\begin{proof}
  The corollary immediately follows from Proposition~\ref{prop:sat}
  and the observation that the cardinality of $\ATab{\AlgS}[n]$ is at
  most one at root~$n$, by properties of the algorithm~$\AlgS$ and
  since $\chi(n) = \emptyset$.
\end{proof}

The following lemma establishes that the \PROJ-solutions up to
root~$n$ of a given tree decomposition solve the \PMC problem.

\begin{lemma}\label{lem:global}
  The
  value~$c = \sum_{\langle\hat\sigma\rangle\text{ is a \PROJ-solution
      up to } n}\Card{I_P(\hat \sigma)}$ if and only if $c$ is the
  projected model count of~$F$ with respect to the set~$P$ of
  projection variables.
\end{lemma}
\begin{proof}
  (``$\Longrightarrow$''): Assume
  that~$c = \sum_{\langle\hat\sigma\rangle\text{ is a \PROJ-solution
      up to } n}\Card{I_P(\hat \sigma)}$. Observe that there can be at
  most one projected solution up to~$n$,
  since~$\chi(n)=\emptyset$. %
  If~$c=0$, then $\ATab{\AlgS}[n]$ contains no rows. Hence, $F$ has no
  models,~c.f., Proposition~\ref{prop:sat}, and obviously also no
  models projected to~$P$. Consequently, $c$ is the projected model
  count of~$F$.  
  If~$c>0$ we have by Corollary~\ref{cor:psat} that~$c$ is
  equivalent to the projected model count of~$F$ with respect to~$P$.

  (``$\Longleftarrow$''): The proof proceeds similar to the only-if
  direction.
\end{proof}

\medskip %

In the following, we provide for a given node~$t$ and a given \PROJ-solution up to~$t$,
the definition of a \PROJ-row solution at~$t$.

\begin{definition}\label{def:loctab}~
  \begin{enumerate}
    \item 
  Let $\hat\sigma$ be a~\PROJ-solution up to some node~$t'$. 
  %
  %
  Then, we define \emph{the local table for node}~$t$ as
  $\local(t,\hat\sigma)\eqdef \{ \langle \vec{\tabval}\rangle \mid
  \langle t, \vec{\tabval}\rangle \in \hat\sigma\}$.

  \item
  Let $t \in N$ be a node of the tree decomposition~$\TTT$ and
  $\langle \hat\sigma \rangle$ be a~$\PROJ$-solution up to~$t$. Then, we
  define the \emph{$\PROJ$-row solution at $t$} by
  $\langle \local(t,\hat\sigma), \Card{I_P(\hat\sigma)}\rangle$.
\end{enumerate}
\end{definition}






\begin{observation}\label{obs:unique}
  Let $\langle \hat \sigma\rangle$ be a \PROJ-solution up to a
  node~$t\in N$.  There is exactly one corresponding \PROJ-row
  solution
  $\langle \local(t,\hat\sigma), \Card{I_P(\hat\sigma)}\rangle$ at~$t$.

  Vice versa, let $\langle \sigma, c\rangle$ at~$t$ be a \PROJ-row
  solution at~$t$ for some integer~$c$. Then, there is exactly one
  corresponding \PROJ-solution~$\langle\PExt_{\leq t}(\sigma)\rangle$
  up to~$t$.
\end{observation}

We need to ensure that storing~$\PROJ$-row solutions at a
node~$t \in N$ suffices to solve the~\PMC problem, which is necessary
to obtain the runtime bounds as presented in
Corollary~\ref{cor:runtime}.

\begin{lemma}\label{lem:local}
  Let $t\in N$ be a node of the tree decomposition~$\TTT$.  There is a
  \PROJ-row solution at the root~$n$ if and only if the projected
  model count of~$F$ is larger than zero.
\end{lemma}
\begin{proof}%

  (``$\Longrightarrow$''): Let $\langle \sigma, c\rangle$ be a
  \PROJ-row solution at root~$n$ where $\sigma$ is a $\AlgS$-table and
  $c$ is a positive integer. Then, by Definition~\ref{def:loctab}
  there also exists a
  corresponding~$\PROJ$-solution~$\langle \hat\sigma \rangle$ up
  to~$n$ such that $\sigma = \local(t,\hat\sigma)$ and
  $c=\Card{I_P(\hat\sigma)}$.
  Moreover, since~$\chi(n)=\emptyset$, we
  have~$\Card{\ATab{\AlgS}[n]}=1$.  
  Then, by Definition~\ref{def:globalsol}
  $\hat \sigma = \ATab{\AlgS}[n]$. By Corollary~\ref{cor:psat}, we
  have $c=\Card{I_P(\ATab{\AlgS}[n])}$.
  Finally, the claim follows.
  

  (``$\Longleftarrow$''): The proof proceeds similar to the only-if
  direction.
\end{proof}

\begin{observation}\label{obs:main_incl_excl}
  Let $n$ be a positive integer, $X = \{1, \ldots, n\}$, and $X_1$,
  $X_2$, $\ldots$, $X_n$ subsets of $X$.
  The number of elements in the intersection over all sets~$A_i$ is
  \begin{align*}
    \Card{\bigcap_{i \in X} X_i} 
    =& %
       \Bigg|\Card{\bigcup^n_{j = 1} X_j} &&- %
                                       \sum_{\emptyset \subsetneq I \subsetneq X, \Card{I}=1}
                                       \Card{\bigcap_{i \in I} X_i} + %
                                       \sum_{\emptyset \subsetneq I \subsetneq X, \Card{I}=2}
                                       \Card{\bigcap_{i \in I} X_i} - \ldots \\
                                         & &&+ \sum_{\emptyset \subsetneq I \subsetneq X, \Card{I}=n-1} (-1)^{\Card{I}} 
                                              \Card{\bigcap_{i \in I} X_i}\Bigg|.
  \end{align*}
It trivially works to count arbitrary sets.
\end{observation}
\begin{proof}
  We take the well-known inclusion-exclusion
  principle~\cite{GrahamGrotschelLovasz95a} and rearrange the
  equation.
  \begin{align*}
    \Card{\bigcup^n_{j = 1} X_j} =& \sum_{\emptyset \subsetneq I \subseteq X} (-1)^{\Card{I}-1} &&\Card{\bigcap_{i \in I} X_i}\\
    \Card{\bigcup^n_{j = 1} X_j} =& \sum_{\emptyset \subsetneq I \subsetneq X} (-1)^{\Card{I}-1} &&\Card{\bigcap_{i \in I} X_i} + (-1)^{\Card{X}-1} \Card{\bigcap_{i \in X} X_i}\\
    (-1)^{\Card{X}-1} \Card{\bigcap_{i \in X} X_i}  =& &&\Card{\bigcup^n_{j = 1} X_j} - \sum_{\emptyset \subsetneq I \subsetneq X} (-1)^{\Card{I}-1} \Card{\bigcap_{i \in I} X_i} \\
    \Card{\bigcap_{i \in X} X_i}  =& \Bigg|&&\Card{\bigcup^n_{j = 1} X_j} - \sum_{\emptyset \subsetneq I \subsetneq X} (-1)^{\Card{I}-1} \Card{\bigcap_{i \in I} X_i}\Bigg| \\
    \Card{\bigcap_{i \in X} X_i} =& \Bigg|%
                                      &&\Card{\bigcup^n_{j = 1} X_j} - %
                                      \sum_{\emptyset \subsetneq I
                                        \subsetneq X, \Card{I}=1}
                                      \Card{\bigcap_{i \in I} X_i} \\ %
    &&&                                 + \sum_{\emptyset \subsetneq I
                                        \subsetneq X, \Card{I}=2}
                                      \Card{\bigcap_{i \in I} X_i} \\
    &&&                                  - \ldots \\
    &&& + \sum_{\emptyset \subsetneq I \subsetneq X,
                                        \Card{I}=n-1} (-1)^{\Card{I}
                                        } \Card{\bigcap_{i \in I}
                                        X_i}\Bigg|
  \end{align*}
\end{proof}

\begin{lemma}\label{lem:main_incl_excl}
  Let $t\in N$ be a node of the tree decomposition~$\TTT$
  with~$\children(t,T) = \langle t_1, \ldots, t_\ell \rangle$ and let
  $\langle\sigma,\cdot\rangle$ be a~\PROJ-row solution at~$t$.
  Then,
  \begin{enumerate}
  \item %
    $\ipmc(t,\sigma,\langle\ATab{\PROJ}[t_1], \ldots,
    \ATab{\PROJ}[t_{\ell}]\rangle) = \ipmc_{\leq t}(\sigma)$
  \item \smallskip%
    for $\type(t) \neq \leaf$:\\
    $\pmc(t,\sigma,\langle\ATab{\PROJ}[t_1], \ldots,
    \ATab{\PROJ}[t_{\ell}]\rangle) = \pmc_{\leq t}(\sigma)$.
  \end{enumerate}
\end{lemma}
\begin{proof}[Sketch]
  We prove the statement by simultaneous induction.
  
  (``Induction Hypothesis''): Lemma~\ref{lem:main_incl_excl} holds for the nodes in~$\children(t,T)$ and also for node~$t$, but on strict subsets~$\rho\subsetneq\sigma$.

  (``Base Cases''): Let $\type(t) = \leaf$.
  Then by definition,
  $\ipmc(t,\emptyset, \langle \rangle) = \ipmc_{\leq t}(\emptyset) =
  1$.  
  Recall that for $\pmc$ the equivalence does not hold for leaves, but we use a node
  that has a node~$t'\in N$ with~$\type(t') = \leaf$ as child for the
  base case. Observe that by definition of a nice tree decomposition
  such a node~$t$ can have exactly one child.
  Then, we have that
  $\pmc(t,\sigma,\langle\ATab{\PROJ}[t']\rangle) = \sum_{\emptyset
    \subsetneq O \subseteq {\origs(t,\sigma)}} (-1)^{(\Card{O} - 1)}
  \cdot \sipmc(\langle \ATab{\AlgS}[t']\rangle, O) =
  \Card{\bigcup_{\vec u\in\sigma} I_P(\PExt_{\leq t}(\{\vec u\}))} =
  \pmc_{\leq t}(\sigma) = 1$ where $\langle\sigma,\cdot\rangle$ is
  a~\PROJ-row solution at~$t$.

  (``Induction Step''): We proceed by case distinction.

  Assume that $\type(t) = \intr$.
  Let $a \in (\chi(t) \setminus \chi(t'))$ be an introduced
  variable. We have two cases. Assume Case (i): $a$ also belongs to
  $(\var(F) \setminus P)$,~i.e., $a$ is not a projection variable. 
  %
  %
  %
  Let~$\langle \sigma, c \rangle$ be a \PROJ-row solution at~$t$ for
  some integer~$c$. By construction of the table algorithm~$\AlgS$
  there are many rows in the table~$\ATab{\AlgS}[t]$ for one row in
  the table~$\ATab{\AlgS}[t']$, more precisely,
  $\Card{\buckets_P(\sigma)} = 1$.
  As a result,
  $\pmc_{\leq t}(\sigma) = \pmc_{\leq t'}(\orig(t,\sigma))$ by
  applying Observation~\ref{obs:unique}.
  We apply the inclusion-exclusion principle on every subset~$\rho$ of
  the origins of~$\sigma$ in the definition of~$\pmc$ and by induction
  hypothesis we know that
  $\ipmc(t',\rho,\langle\ATab{\PROJ}[t']\rangle) = \ipmc_{\leq
    t'}(\rho)$, therefore,
  $\sipmc(\ATab{\PROJ}[t'], \rho) = \ipmc_{\leq t'}(\rho)$.  This
  concludes Case~(i) for $\pmc$. The induction step for $\ipmc$ works
  similar, but swapped by applying
  Observation~\ref{obs:main_incl_excl} and comparing the corresponding
  \PROJ-solutions up to~$t$ or $t'$, respectively. Further, for showing the lemma for~$\ipmc$, one has to additionally apply the hypothesis for node~$t$, but on strict subsets~$\emptyset\subsetneq\rho\subsetneq\sigma$ of~$\sigma$.
  %
  Assume that we have Case~(ii): $a$ also belongs to $P$,~i.e., $a$ is a projection
  variable. We proceed similar as in Case~(i), and obtain that
  $\Card{\buckets_P(\sigma)} = 1$.

  Assume that $\type(t) = \rem$. Let
  $a \in (\chi(t') \setminus \chi(t))$ be a removed variable. We have
  two cases. Case (i) $a$ also belongs to
  $(\var(F) \setminus P)$,~i.e., $a$ is not a projection variable; and
  Case (ii) $a$ also belongs to $P$,~i.e., $a$ is a projection
  variable.
  Assume that we have Case~(i).  Let~$\langle \sigma, c \rangle$ be a
  \PROJ-row solution at~$t$ for some integer~$c$.
  By construction of the table algorithm~$\AlgS$ there are many rows
  in the table~$\ATab{\AlgS}[t]$ for one row in the
  table~$\ATab{\AlgS}[t']$ (and vice-versa). Nonetheless we still have
  $\pmc_{\leq t}(\sigma) = \pmc_{\leq t'}(\orig(t,\sigma))$, because
  $a \notin P$ by applying Observation~\ref{obs:unique}.
  We apply the inclusion-exclusion principle on every subset~$\rho$ of
  the origins of~$\sigma$ in the definition of~$\pmc$ and by induction
  hypothesis we know that
  $\ipmc(t',\rho,\langle\ATab{\PROJ}[t']\rangle) = \ipmc_{\leq
    t'}(\rho)$, therefore,
  $\sipmc(\ATab{\PROJ}[t'], \rho) = \ipmc_{\leq t'}(\rho)$.  This
  concludes Case~(i) for $\pmc$. Again, the induction step for $\ipmc$
  works similar, but swapped.
  Assume that we have Case~(ii).
  Let~$\langle \sigma, c \rangle$ be a \PROJ-row solution at~$t$ for
  some integer~$c$.
  Here we cannot ensure
  $\pmc_{\leq t}(\sigma) = \pmc_{\leq t'}(\orig(t,\sigma))$, since
  buckets fall together.  However, by applying
  Observation~\ref{obs:unique} we have
  $\pmc_{\leq t}(\sigma) = \sum_{\rho \in
    \buckets_P(\origs(t,\sigma)_{(1)})} \pmc(t', \rho, C) $ where the
  sequence~$C$ consists of the tables of the children of~$t'$.
  For every~$\rho \in \subbuckets_P(\origs(t,\sigma)_{(1)})$ by
  induction hypothesis we know that
  $\ipmc(t',\rho,\langle\ATab{\PROJ}[t']\rangle) = \ipmc_{\leq
    t'}(\rho)$.
  Hence, we apply the inclusion-exclusion principle over all
  subsets~$\zeta$ of~$\rho$ for all~$\rho$ independently.  By
  construction
  $\sipmc(\ATab{\PROJ}[t'], \zeta) = \ipmc_{\leq t'}(\zeta)$.  Then,
  by construction
  $\pcnt(t,\sigma, C') = \sum_{\emptyset \subsetneq O \subseteq
    {\origs(t,\sigma)}} (-1)^{(\Card{O} - 1)} \cdot \sipmc(C', O) =
  \pmc_{\leq t}(\sigma)$ where
  $C' = \langle \ATab{\PROJ}[t'] \rangle$, since for the remaining
  terms $\sipmc(C', O)$ is simply zero, including cases where
  different buckets are involved.
  This concludes Case~(ii) for $\pmc$. Again, the induction step for
  $\ipmc$ works similar, but swapped by again applying
  Observation~\ref{obs:main_incl_excl}.

  Assume that $\type(t) = \join$. We proceed similar to the introduce
  case. However, we have two \PROJ-tables for the children of~$t$.
  Hence, we have to both sides when computing $\sipmc$
  (Definition~\ref{def:childpcnt}). There we consider the
  cross-product of two \AlgS-tables and we can also correctly apply
  the inclusion-exclusion principle on subsets of this cross-product,
  which we can do by simply multiplying $\sipmc$-values
  accordingly. The multiplication is closely related to the join case
  in table algorithm~\AlgS. For $\ipmc$ this does not apply, since the
  inclusion-exclusion principle is carried out at the node~$t$ and not
  for its children.

  Since we outlined all cases that can occur for node~$t$, this
  concludes the proof sketch.
\end{proof}


\begin{lemma}[Soundness]\label{lem:correct}
  Let $t\in N$ be a node of the tree decomposition~$\TTT$
  with~$\children(t,T) = \langle t_1, \ldots, t_\ell \rangle$.
  Then, each row~$\langle \tab{}, c \rangle$ at node~$t$ constructed
  by table algorithm~$\PROJ$ is also a~\PROJ-row solution for
  node~$t$.
\end{lemma}
\begin{proof}[Idea]
  Observe that Listing~\ref{fig:dpontd3} computes a row for each
  sub-bucket $\sigma \in \subbuckets_P(\ATab{\AlgS}[t])$. The
  resulting row~$\langle\sigma, c \rangle$ obtained by~$\ipmc$ is
  indeed a \PROJ-row solution for~$t$ according to
  Lemma~\ref{lem:main_incl_excl}.
\end{proof}


\begin{lemma}[Completeness]\label{lem:complete}
  Let~$t\in N$ be node of the tree decomposition~$\TTT$ where
  $\type(t) \neq \leaf$.  Given a
  \PROJ-row solution~$\langle \sigma, c \rangle$ at node~$t$.
  There exists $\langle C_1, \ldots, C_\ell\rangle$ where $C_i$ is set
  of \PROJ-row solutions of the from~$\langle\sigma_i, c_i\rangle$
  such that
  $\sigma = \PROJ(t, \cdot, \cdot, P, \langle C_1, \ldots,
  C_\ell\rangle, \ATab{\AlgS})$.
\end{lemma}
\begin{proof}[Idea]
Since~$\langle\sigma,c \rangle$ is a~\PROJ-row solution for~$t$, there is by Definition~\ref{def:loctab} a corresponding ~\PROJ-solution~$\langle\hat\sigma\rangle$ up to~$t$ such that~$\local(t,\hat\sigma) = \sigma$. 

We proceed again by case distinction. Assume that~$\type(t)=\intr$. Then we define~$\hat{\sigma'}\eqdef \{(t',\hat\rho) \mid (t', \hat\rho)\in \sigma, t \neq t'\}$. Then, for each subset~$\emptyset\subsetneq\rho\subseteq\local(t',\hat{\sigma'})$, we define~$\langle \rho, \Card{I_P(\PExt_{\leq t}(\rho))}\rangle$ in accordance with Definition~\ref{def:loctab}. By Observation~\ref{obs:unique}, we have that~$\langle \rho, \Card{I_P(\PExt_{\leq t}(\rho))}\rangle$ is a \AlgS-row solution at node~$t'$. 
Since we defined the~\PROJ-row solutions for~$t'$ for all the respective \PROJ-solutions up to~$t'$, we encountered every~\PROJ-row solution for~$t'$ that is required for deriving~$\langle \sigma, c\rangle$ via~\PROJ (c.f. Definitions~\ref{def:ipmc} and~\ref{def:pcnt}).

Assume that~$\type(t)=\rem$. The case is slightly easier as the one
above. We do not need to define a~\PROJ-row solution for~$t$' for all
subsets~$\rho$, since we only have to consider subsets~$\rho$ here,
with~$\Card{\buckets_P(\rho)}=1$. The remainder works similar.

Similarly, one can show the result for the remaining node with~$\type(t)=\join$, but define \PROJ-row solutions for two preceding child nodes of~$t$.
\end{proof}

We are now in the position to proof our theorem.

\begin{restatetheorem}[thm:correctness]%
\begin{theorem}
  The algorithm~$\dpa_\PROJ$ is correct. \\
  More precisely, 
  %
  the algorithm~$\dpa_\PROJ((F,P),\TTT,\ATab{\AlgS})$ returns
  tables~$\ATab{\PROJ}$ such that $c=\sipmc(\ATab{\AlgS}[n], \cdot)$
  is the projected model count of~$F$ with respect to the set~$P$ of
  projection variables.
\end{theorem}
\end{restatetheorem}
\begin{proof}
  %
  By Lemma~\ref{lem:correct} we have soundness for every
  node~$t \in N$ and hence only valid rows as output of table
  algorithm~$\PROJ$ when traversing the tree decomposition in
  post-order up to the root~$n$.
  By Lemma~\ref{lem:local} we know that the projected model count~$c$
  of~$F$ is larger than zero if and only if there exists a
  certain~\PROJ-row solution for~$n$.
  This~\PROJ-row solution at node~$n$ is of the
  form~$\langle \{\langle\emptyset, \ldots\rangle\} ,c\rangle$. If
  there is no \PROJ-row solution at node~$n$,
  then~$\ATab{\AlgS}[n]=\emptyset$ since the table algorithm~$\AlgS$
  is correct (c.f. Proposition~\ref{prop:sat}). Consequently, we have
  $c=0$. Therefore, $c=\sipmc(\ATab{\AlgS}[n], \cdot)$ is the
  projected model count of~$F$ with respect to~$P$ in both cases.
  %
  %
  
  %
  %

  %
  %

  Next, we establish completeness by induction starting from the
  root~$n$. Let therefore, $\langle \hat\sigma \rangle$ be the~\PROJ-solution up to node~$n$, where for each row
  in~$\vec u\in \hat\sigma$, $I(\vec u)$ corresponds to a model of~$F$.  By
  Definition~\ref{def:loctab}, we know that for the root~$n$ we
  can construct a \PROJ-row solution at~$n$ of the
  form~$\langle \{\langle\emptyset, \ldots\rangle\} ,c\rangle$
  for~$\hat\sigma$.  We already established the induction step in
  Lemma~\ref{lem:complete}.
  Hence, we obtain some (corresponding) rows for every
  node~$t$. Finally, we stop at the leaves.

  In consequence, we have shown both soundness and completeness. As a
  result, Theorem~\ref{thm:correctness} is sustains.
\end{proof}

\begin{restatecorollary}[cor:correctness]%
\begin{corollary}
  The algorithm $\mdpa{\AlgS}$ is correct and outputs for any instance
  of \PMC its projected model count.
\end{corollary}
\end{restatecorollary}
\begin{proof}
  The result follows immediately, since~$\mdpa{\AlgS}$ consists of two
  dynamic programming passes~$\dpa_\AlgS$, a purging step and~$\dpa_\PROJ$. For the
  soundness and completeness of~$\dpa_\AlgS$ we refer to other
  sources~\cite{SamerSzeider10b,FichteEtAl17a}. By Proposition~\ref{prop:sat}, the
  ``purging'' step does neither destroy soundness nor completeness
  of~$\dpa_\PRIM$.
\end{proof}

\begin{corollary}
  Unless ETH fails, $\PMC$ cannot be solved in
  time~$2^{2^{o(k)}}\cdot \CCard{F}^{o(k)}$ for a given instance
  $(F,P)$ where~$k$ is the treewidth of the incidence graph of~$F$.
\end{corollary}
\begin{proof}
  Let $w_i$ and $w_p$ be the treewidth of the incidence graph and
  primal graph of~$F$, respectively. Then,
  $w_i \leq w_p +1$~\cite{SamerSzeider10b}, which establishes the
  claim.
\end{proof}

}

\end{document}